\documentclass[10pt]{article}
\usepackage{bigai_preprint}
\usepackage{amsmath,amsfonts,bm}

\def\eqref#1{equation~\ref{#1}}

\def\1{\bm{1}}

\DeclareMathAlphabet{\mathsfit}{\encodingdefault}{\sfdefault}{m}{sl}
\SetMathAlphabet{\mathsfit}{bold}{\encodingdefault}{\sfdefault}{bx}{n}

\usepackage{needspace}
\usepackage{threeparttable}

\usepackage{xspace}
\usepackage{balance}
\usepackage{setspace}
\usepackage{etoolbox}
\usepackage{nicefrac}                 %

\AtBeginDocument{
    \setlength{\parskip}{4pt}
}

\titlespacing*{\paragraph}{0pt}{0.6ex plus 0.2ex minus .1ex}{0.65em}
\usepackage{tabularx}
\usepackage{multirow,array,makecell}
\usepackage{colortbl}                 %
\usepackage{longtable}
\usepackage{siunitx}
\usepackage{rotating}
\usepackage{adjustbox}
\usepackage{wrapfig}
\usepackage{overpic}
\usepackage{hvfloat}
\usepackage{subcaption}               %
\usepackage{float}
\usepackage{placeins}               %
\usepackage{pifont}
\usepackage[misc]{ifsym}
\usepackage{xcolor}

\definecolor{uclablue}{RGB}{39,116,174}
\definecolor{taskred}{RGB}{190,55,55}

\usepackage[linesnumbered,ruled,vlined]{algorithm2e}
\usepackage{algorithmic}

\usepackage{acronym}
\acrodef{mle}[MLE]{Maximum Likelihood Estimation}
\acrodef{maxent}[Max-Ent]{Maximum Entropy}
\acrodef{merl}[Max-Ent RL]{Maximum-Entropy Reinforcement Learning}
\acrodef{hoi}[HOI]{Human-Object Interaction}
\acrodef{nerf}[NeRF]{Neural Radiance Field}
\acrodef{mdl}[MDL]{Minimum Description Length}
\acrodef{em}[EM]{Expectation-Maximization}

\makeatletter
\DeclareRobustCommand\onedot{\futurelet\@let@token\@onedot}
\def\@onedot{\ifx\@let@token.\else.\null\fi\xspace}

\makeatother

\newcommand{\cmark}{\ding{51}}
\newcommand{\xmark}{\ding{55}}

\newcommand{\method}{\textbf{Zero-HLM}\xspace}

\newcommand{\compacttable}{%
    }

\usepackage[capitalise]{cleveref}   
\crefname{algorithm}{Alg.}{Algs.}
\Crefname{algocf}{Algorithm}{Algorithms}
\crefname{section}{Sec.}{Secs.}
\Crefname{section}{Section}{Sections}
\crefname{table}{Tab.}{Tabs.}
\Crefname{table}{Tab.}{Tabs.}
\crefname{figure}{Fig.}{Figs.}
\Crefname{figure}{Fig.}{Figs.}
\crefname{equation}{Eq.}{Eqs.}
\Crefname{equation}{Equation}{Equations}

\newcommand{\papertitle}{Zero-HLM: Generalizable Humanoid Loco-Manipulation without Real Robot Data}

\newcommand{\papertitles}{\papertitle}

\newcommand{\paperauthors}{%
Dianyi Yang\textsuperscript{1,2,*}, Mengya Liu\textsuperscript{2,*},
Zhumei Wang\textsuperscript{1,2}, Yu Liu\textsuperscript{2,3},
Xiaochen Zhang\textsuperscript{1,2}, Yixuan Li\textsuperscript{1},\\[2pt]
Le Ma\textsuperscript{2}, Yuhang Yang\textsuperscript{2,3},
Weiqi Huang\textsuperscript{1,2}, Wei Liang\textsuperscript{1,\textdagger},
Baoxiong Jia\textsuperscript{2,\textdagger,\textdaggerdbl}, Siyuan Huang\textsuperscript{2,4,\textdagger}}
\newcommand{\paperaffiliations}{%
\textsuperscript{1}\,Beijing Institute of Technology\quad
\textsuperscript{3}\,Tsinghua University\quad
\textsuperscript{4}\,Delta Intelligence\\
\textsuperscript{2}\,State Key Laboratory of General Artificial Intelligence, BIGAI\\[3pt]
{\footnotesize\normalfont
\textsuperscript{*}\,Equal contribution\quad
\textsuperscript{\textdagger}\,Corresponding authors\quad\textsuperscript{\textdaggerdbl}\,Project Lead}\\[2pt]
{\footnotesize\normalfont\bfseries Project page: \href{https://young-bit.github.io/zero-hlm/}{\texttt{https://young-bit.github.io/zero-hlm/}}}}

    \title{\parbox{0.96\textwidth}{\centering
        \papertitles}}
    \shorttitle{\papertitle}
    \author{\paperauthors}
    \affiliations{\paperaffiliations}
    \paperdate{Technical Report}
    \headerlogo{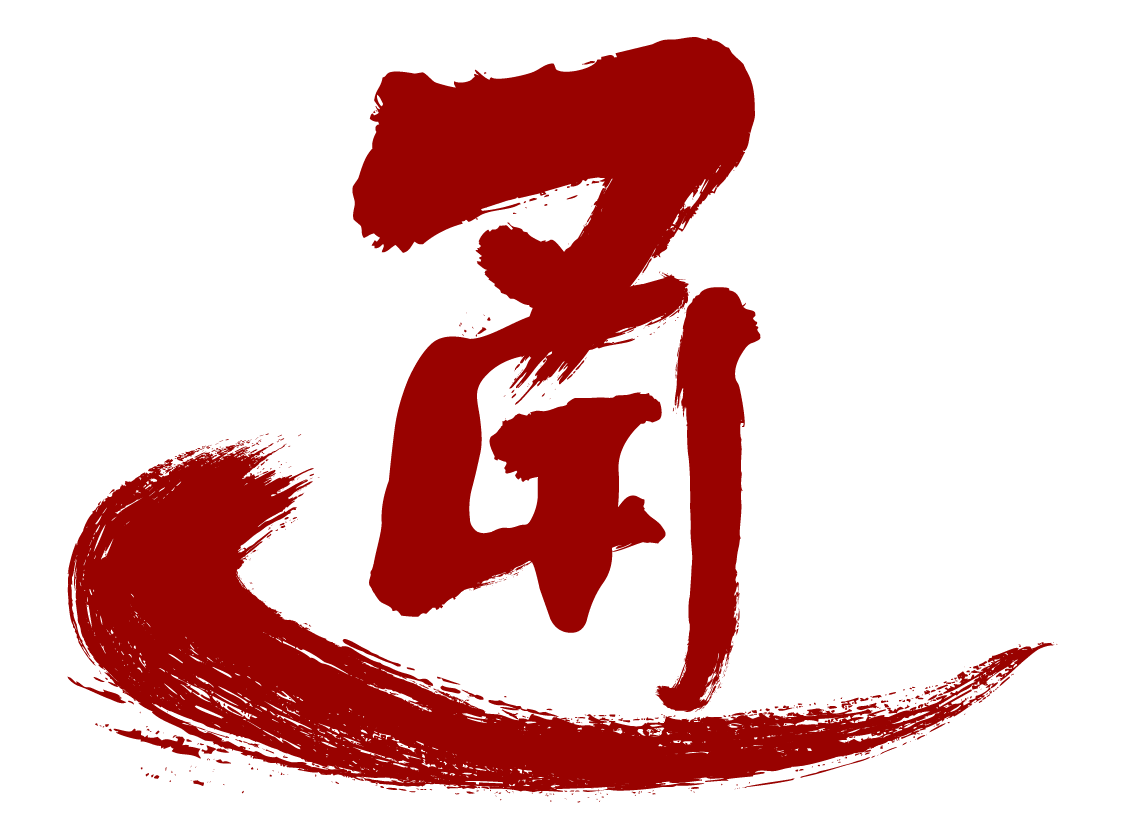}
    \abstractlogo{figures/logo.png}

\newcommand{\paperbibstyle}{unsrtnat}
\begin{document}
\maketitle
\begingroup
\setlength{\intextsep}{6pt}
\begin{figure}[H]
    \centering
    \captionsetup{skip=4pt}
    
    \includegraphics[width=\textwidth,height=0.40\textheight,keepaspectratio]{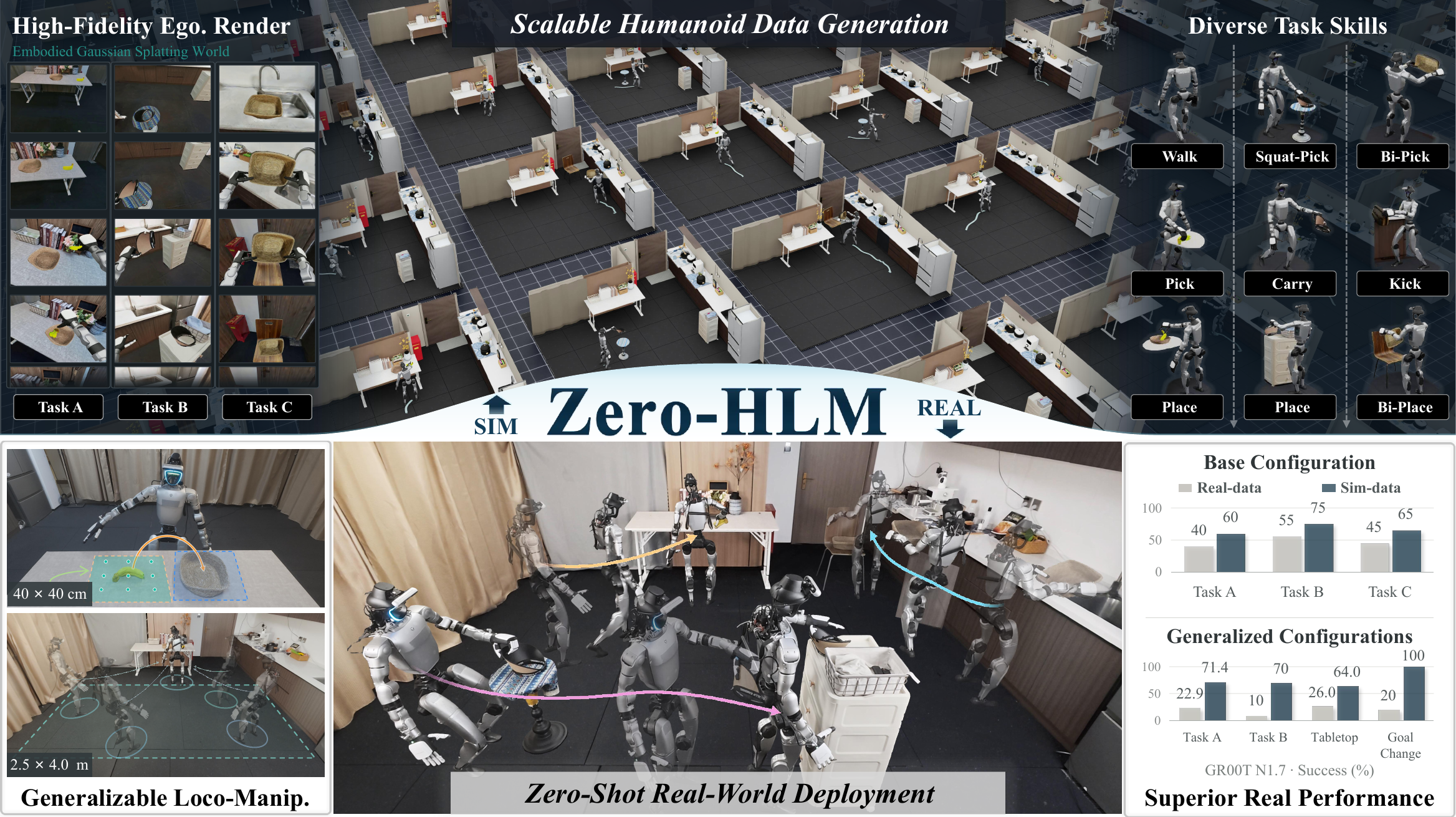}
    
    \caption{\textit{Top:}  \method generates humanoid loco-manipulation data covering diverse whole-body skills in simulation. \textit{Bottom:} Policies trained solely on these data achieve direct sim-to-real transfer on a Unitree G1 under \textsc{Base} and generalize across \textsc{Expanded} configurations.}
    \label{fig:teaser}
\end{figure}
\endgroup
\begin{abstract}

Learning generalizable humanoid loco-manipulation remains bottlenecked by costly real-world demonstrations. Simulation offers a scalable alternative, but extending it to humanoids requires solving two coupled challenges: constructing scene-scale environments that are both visually faithful and physics-ready, and generating successful demonstrations under unified whole-body control. We present \method, a real-to-sim-to-real framework that transforms smartphone captures and simulated source demonstrations into scalable training data for humanoid loco-manipulation. We couple Gaussian visual reconstruction with agentic physical scene construction to build interaction-ready digital twins with aligned visual and physical geometry. Within the environment, our motion--environment co-adaptation strategy synthesizes unified whole-body references and uses rollout feedback to restore interaction geometry when tracking deviations disrupt execution. From 45 source demonstrations teleoperated in simulation, \method generates \textbf{4,800} physically validated demonstrations, averaging a \textbf{1.8}$\boldsymbol{\times}$ speedup over real-world data collection. Policies trained solely on these data transfer zero-shot to a Unitree G1 across three long-horizon tasks. With GR00T N1.7, simulation-only training achieves \textbf{66.7\%} mean full-task success with a \textbf{20-percentage-point} gain over real-data training under matched configurations. Across five generalization settings, pooled real-world success increases from \textbf{23\%} to \textbf{69\%} without additional real demonstrations. To support systematic evaluation, we establish a benchmark with nine skill stages and shared protocols for comparing models and studying data scaling.
\end{abstract}

\clearpage
\section{Introduction}
\label{sec:intro}

Humanoid robots combine mobility and dexterity with a morphology naturally suited to human environments. Recent advances in whole-body control and visuomotor policies have enabled diverse loco-manipulation behaviors~\citep{luo2026sonic,wu2026teleopit,nvidia2025grootn1,black2024pi0,li2026omega0,jiang2025wholebodyvla,shi2026egohumanoid,fu2024humanplus}. 
However, learning generalizable behaviors requires diverse interaction data that remain costly and difficult to collect on a physical humanoid~\citep{yu2026oasis,kim2026legs,zhong2025humanoidexo,seo2023trill}. 

Simulation-based data generation offers a scalable alternative to costly physical collection, but its practical value depends on constructing simulation environments that support reliable sim-to-real transfer~\citep{xiang2026functionpreserving}. For humanoids, locomotion expands the workspace beyond a local manipulation area, while a moving egocentric camera observes the environment from changing positions and heights. This calls for scene-level reconstruction that preserves both visual fidelity and the physical geometry and dynamics required for interaction. Existing Gaussian-based real-to-sim-to-real methods have shown promising transfer in tabletop settings~\citep{yang2025robosplat,han2026re3sim,yu2025real2render2real,qureshi2025splatsim,barcellona2025dream,li2024robogsim,jiang2026gsworld}. However, extending this paradigm to scene-scale humanoid interaction is considerably more demanding, as meshes extracted from Gaussian reconstructions can contain contact artifacts, and articulated mechanisms require additional modeling effort~\citep{han2026re3sim,lee2026rora}.

Beyond reconstructing such environments, generating humanoid interaction data at scale remains challenging. Recent methods~\citep{kim2026legs,lin2026humanoidmimicgen,fu2025demohlm} simplify data generation by specifying locomotion and upper-body commands
separately. This separation can limit motion naturalness and support for
leg-critical interactions, motivating unified whole-body reference motion tracking~\citep{luo2026sonic,wu2026teleopit}.
However, scaling data generation under unified tracking
presents two challenges: (i) \textit{Whole-body reference motion synthesis}. Adapting source demonstrations to varied spatial configurations requires coordinated changes throughout the body while preserving task-relative interaction~\citep{yang2026omniretarget}. (ii) \textit{Execution mismatch}. Whole-body controllers must reconcile reference tracking with balance and contact constraints, so precise state-by-state replay cannot be assumed~\citep{liao2026beyondmimic}. Even kinematically valid reference motions may produce tracking deviations in the robot base and hands, disrupting intended object interactions.

To address these challenges, we propose Zero-HLM, a real-to-sim-to-real framework that combines visually faithful, physics-ready real-to-sim reconstruction with scalable demonstration generation under unified whole-body control.
First, inspired by recent MLLM-based agentic frameworks that produce faithful digital cousin reconstructions~\citep{dai2024automated,yu2025metascenes}, we propose \textbf{Gaussian-Guided Agentic Real-to-Sim}, which builds a scene-level digital twin by coupling a high-fidelity Gaussian reconstruction with an agentic pipeline that iteratively aligns the digital cousins to the real scene, using the Gaussian reconstruction as the alignment reference. We further composite the Gaussian rendering with the simulator's rendering of the aligned scene, yielding a hybrid rendering that narrows the visual sim-to-real gap.
To tackle the mismatch between synthesized reference motions and their tracked execution, we introduce a motion--environment co-adaptation strategy for scalable data generation under unified whole-body control. \textbf{Demonstration-Anchored Whole-Body Motion Synthesis} combines locomotion planning with object-relative adaptation of source demonstrations, producing continuous whole-body reference motions for varied spatial configurations. \textbf{Execution-Conditioned Correction} then uses rollout feedback to adjust scene configurations and reference motions, restoring interactions disrupted by tracking deviations. We pair the resulting physically validated trajectories with photorealistic egocentric observations for policy training. Policies trained solely on our simulated data transfer directly to the real world and generalize across diverse scene configurations. Our contributions can be summarized as follows:
\begin{itemize}[leftmargin=*,itemsep=1pt,parsep=0pt]
    \item We propose \method, a real-to-sim-to-real framework
    for unified humanoid loco-manipulation. It combines
    Gaussian-guided agentic reconstruction with whole-body
    data generation in visually faithful, physics-ready scenes.

    \item We introduce \textbf{motion--environment co-adaptation}, which couples demonstration-anchored motion synthesis with execution-conditioned correction to compensate for tracking deviations, yielding \textbf{4,800} successful trajectories across three tasks at a \textbf{1.8$\times$} speedup over real-world collection.

    \item We demonstrate \textbf{direct sim-to-real transfer
    and real-world generalization} through simulation-only training.
    With GR00T N1.7, simulation-only training yields a \textbf{20-percentage-point gain} over real-data training under matched configurations. Across five extended generalization settings, pooled real-world success increases from \textbf{23\% to 69\%} without additional real demonstrations.

    \item We establish a \textbf{humanoid loco-manipulation benchmark}
    comprising three long-horizon tasks and nine skill stages, 
    enabling systematic comparisons across diverse policy families under unified training and execution protocols.
\end{itemize}

\clearpage
\begin{table}[H]
\centering
\captionsetup{position=top,belowskip=0pt,skip=2pt}
\caption{
Comparison of humanoid loco-manipulation data-generation methods.
\textbf{Unified Policy}: unified whole-body action prediction.
\textbf{Physics-Ready Real-to-Sim}: automated scene-level reconstruction with articulated components for physical simulation.
\textbf{Few-Demo Skill Adapt.}: task-specific skill adaptation from few source demonstrations.
\textbf{Leg-Critical Interaction}: direct leg/foot--object interaction, excluding locomotion and squatting.
}
\label{tab:humanoid_data_comparison}

\vspace{2pt}
\renewcommand{\arraystretch}{1.16}
\setlength{\tabcolsep}{5.0pt}
\small

\compacttable

\begin{adjustbox}{max width=\linewidth}
\begin{tabular}{@{}lccccccc@{}}
\toprule

&
\multicolumn{1}{c}{\textbf{Policy}} &
\multicolumn{3}{c}{\textbf{Data Generation Pipeline}} &
\multicolumn{3}{c}{\textbf{Capability \& Deployment}} \\

\cmidrule(l{1pt}r{2pt}){2-2}
\cmidrule(l{2pt}r{2pt}){3-5}
\cmidrule(l{2pt}r{0pt}){6-8}

\textbf{Method}
& \makecell{\textbf{Unified}\\\textbf{Policy}}
& \makecell{\textbf{Physics-Ready}\\\textbf{Real-to-Sim}}
& \makecell{\textbf{Few-Demo}\\\textbf{Skill Adapt.}}
& \makecell{\textbf{Photorealistic}\\\textbf{Render}}
& \makecell{\textbf{Articulated}\\\textbf{Task}}
& \makecell{\textbf{Leg-Critical}\\\textbf{Interaction}}
& \makecell{\textbf{Zero-Shot}\\\textbf{Deployment}} \\

\midrule

DemoHLM~\citep{fu2025demohlm}
& \xmark
& \xmark
& \cmark
& \xmark
& \cmark
& \xmark
& \cmark \\

VIRAL~\citep{he2026viral}
& \xmark
& \xmark
& \xmark
& \xmark
& \xmark
& \xmark
& \cmark \\

FetchMan~\citep{rayyan2026fetchman}
& \xmark
& \xmark
& \xmark
& \xmark
& \xmark
& \xmark
& \cmark \\

HumanoidMimicGen~\citep{lin2026humanoidmimicgen}
& \xmark
& \xmark
& \cmark
& \xmark
& \xmark
& \xmark
& \xmark \\

LEGS~\citep{kim2026legs}
& \xmark
& \xmark
& \xmark
& \cmark
& \xmark
& \xmark
& \cmark \\

OASIS~\citep{yu2026oasis}
& \cmark
& \xmark
& \xmark
& \cmark
& \xmark
& \xmark
& \cmark \\

VLK~\citep{wang2026vlk}
& \cmark
& \xmark
& \xmark
& \cmark
& \xmark
& \xmark
& \cmark \\

\midrule

\rowcolor{black!5}
\textbf{Ours}
& \cmark
& \cmark
& \cmark
& \cmark
& \cmark
& \cmark
& \cmark \\

\bottomrule
\end{tabular}
\end{adjustbox}

\end{table}
\vspace{-24pt}

\section{Related Work}
\label{sec:related_work}

\paragraph{Humanoid Loco-Manipulation.}
Humanoid loco-manipulation depends on controllers that coordinate mobility with
contact-rich interaction. Decoupled systems such as HOMIE~\citep{ben2025homie}
combine lower-body control with separately commanded arms and hands. Unified
whole-body trackers, including SONIC~\citep{luo2026sonic} and
OmniClone~\citep{li2026omniclone}, instead track coordinated reference motions
through a shared control interface, enabling human-motion-driven teleoperation
and demonstration collection. Although
teleoperation provides embodiment-aligned interaction data, expanding its task
and scene coverage requires additional operator time and robot rollouts.

Robot-free interfaces such as HuMI~\citep{nai2026humi} and egocentric approaches
such as EgoHumanoid~\citep{shi2026egohumanoid} reduce dependence on a physical
robot during demonstration collection, but still rely on human-recorded
interactions to cover new settings.
In parallel, vision-language-action and world-action models have advanced
generalizable robot learning~\citep{black2025pi05,nvidia2025grootn1,ye2026dreamzero,ma2026dit4dit},
but still depend on diverse interaction data.
In this work, we address this data bottleneck through scalable whole-body
data generation in simulation, enabling zero-shot real-world deployment
with generalization to unseen scene configurations.

\paragraph{Real-to-Sim-to-Real for Robot Learning.}
Real-to-sim-to-real learning leverages reconstructed environments to generate simulation data for real-world policy transfer. Gaussian-based methods preserve captured appearance to reduce the visual sim-to-real gap~\citep{yang2025robosplat,han2026re3sim,qureshi2025splatsim}, while recent agentic approaches facilitate physical scene construction through geometry modeling, physical parameter inference, and simulator configuration~\citep{chen2026agenticreal2sim,li2026arstag}. Our approach brings these capabilities together by grounding agentic scene construction in Gaussian geometry and matched-view renderings, aligning physical interaction geometry with the reconstructed visual scene.

\textit{For demonstration generation}, MimicGen adapts source demonstrations to new object configurations through simulation rollouts~\citep{mandlekar2023mimicgen}, whereas DemoGen performs demonstration and point-cloud transformations without simulation~\citep{xue2025demogen}. Extending data generation to humanoids, VIRAL and FetchMan rely on task-specific rewards~\citep{he2026viral,rayyan2026fetchman}, while other approaches decouple locomotion from upper-body control~\citep{fu2025demohlm,lin2026humanoidmimicgen,kim2026legs}. VLK adopts unified whole-body control but focuses primarily on single-object transport~\citep{wang2026vlk}. In contrast, we adapt source interaction skills without specialized motion generators and correct tracking deviations through motion--environment co-adaptation, validating the resulting motions under the deployment controller. \Cref{tab:humanoid_data_comparison} compares these approaches.

\section{The \method Framework}
\label{sec:method}

\begin{figure*}[t]
    \centering
    \includegraphics[width=\textwidth]{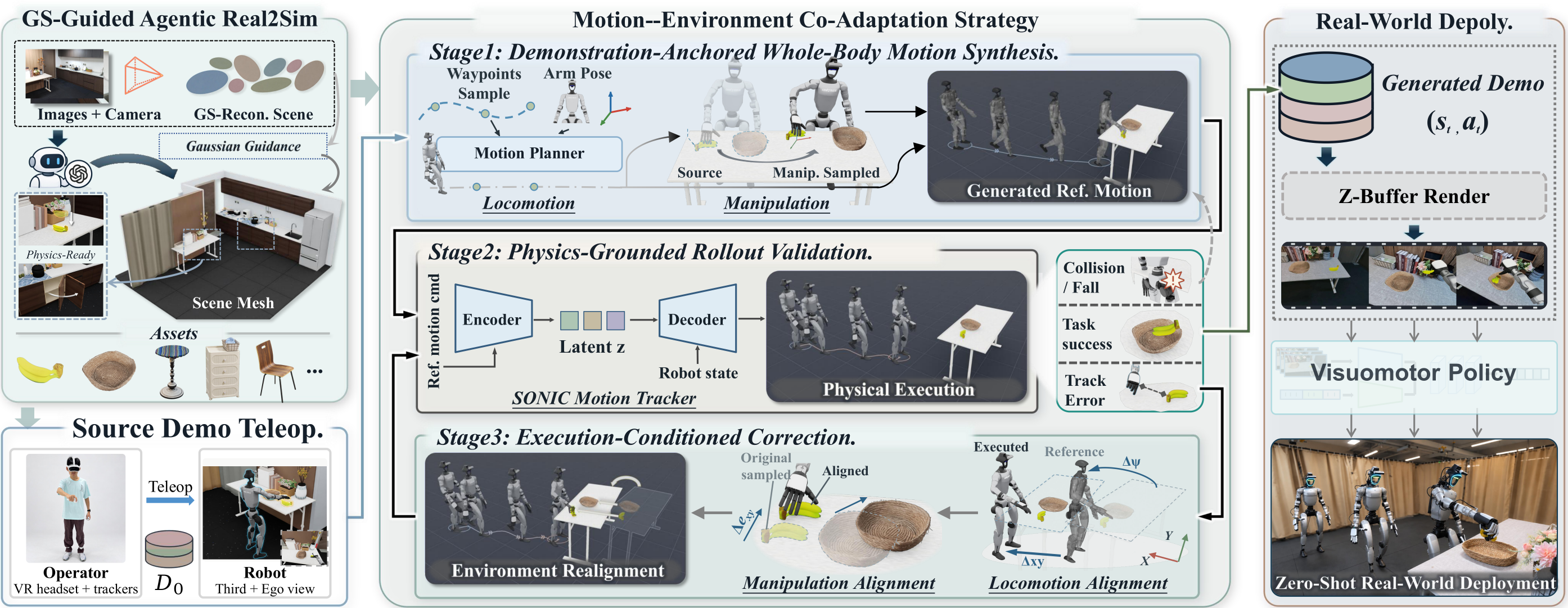}
    \captionsetup{skip=4pt,belowskip=0pt}
    \caption{\textbf{Overview of \method.}
    \textit{Left:} Gaussian-guided agentic real-to-sim reconstruction aligns the physical scene with captured appearance and geometry.
    \textit{Middle:} Motion--environment co-adaptation generates whole-body data from source demonstrations.
    \textit{Right:} successful rollouts are paired with egocentric renderings to train policies for zero-shot real-world deployment.}
    \label{fig:method_overview}
    
\end{figure*}

Our goal is to reconstruct a visually faithful, physics-ready simulation environment from smartphone captures and generate training data at scale for generalizable humanoid loco-manipulation. \Cref{fig:method_overview} illustrates the overall framework. The following sections describe agentic real-to-sim reconstruction (\cref{sec:real2sim}) that supports both physical interaction and egocentric rendering, motion--environment co-adaptation for scalable data generation (\cref{sec:trajectory_generation}), and real-world deployment (\cref{sec:observation_rendering}).

\subsection{Gaussian-Guided Agentic Real-to-Sim Reconstruction}
\label{sec:real2sim}
To narrow the sim-to-real gap, we aim to construct a visually faithful, physics-ready simulation environment with aligned visual and physical representations. We build this environment $\mathcal{E}$ from hybrid representations: a Gaussian reconstruction $\mathcal{G}$ for appearance and spatial reference, and an editable mesh-based physical scene for collision and articulation. We reconstruct $\mathcal{G}$ from smartphone images using MAtCha~\citep{guedon2025matcha}, with checkerboards establishing metric scale and ground alignment.
A multimodal agent then uses $\mathcal{G}$ and a mesh extracted
from it to construct and configure the physical scene using Blender MCP and IsaacLab.
The extracted mesh guides the initial layout and geometry,
while $\mathcal{G}$ provides an active-view rendering interface
for local refinement.
The agent therefore selects inspection viewpoints and compares Gaussian and mesh renderings at matched camera poses to iteratively refine local geometry and placement. The resulting hybrid scene $\mathcal{E}$ integrates a mesh-based physics backend with a Gaussian visual frontend in a shared metric frame, aligning physical interactions with visual observations.
More details appear in Appendix~\ref{app:agentic_real2sim}.

Interaction-critical assets are reconstructed separately from dedicated video captures to preserve the local geometry required for grasping, placement, and physical contact. Camera poses are estimated using COLMAP, while checkerboard calibration establishes metric scale for asset reconstruction. Each asset is then aligned and placed within the scene by the agent according to the task configuration. In IsaacLab, the agent further configures collision geometry, articulated joints, and physical parameters including mass, friction, and center of mass, then refines them through interaction checks. For example, cabinet opening tests articulated motion, while object release tests gravity-driven motion and collision.

\subsection{Motion--Environment Co-Adaptation for Scalable Data Generation}
\label{sec:trajectory_generation}
\label{app:source_annotation}
\label{app:manipulation_reference}
\label{app:manipulation_details}
\label{app:velocity_adaptive_stitching}

In this section, we aim to generate physically executable whole-body demonstrations across diverse spatial configurations from a small set of source demonstrations. We first collect 15 successful simulation teleoperation demonstrations per task as the source pool $\mathcal{D}_0$. Each demonstration is segmented into locomotion, manipulation, and leg-critical interaction using event annotations. For each sampled configuration, we construct a \emph{whole-body reference motion} $\mathcal{R}=(r_1,\ldots,r_T)$ from $\mathcal{D}_0$, where each pose $r_t=(p_t,R_t,q_t)$ specifies the root position, orientation, and body joints. A simulated \emph{rollout} executes this motion through SONIC~\citep{luo2026sonic}, with separate hand commands. Successful rollouts are retained in the dataset $\mathcal{D}$. Here, \emph{stage} denotes a generation step and \emph{segment} a temporal portion of the motion.

\paragraph{Source Annotation and Selection.}
We identify interaction segments using task-specific event predicates based on hand closure, object proximity, displacement, and task success conditions. Only demonstrations with valid interaction events and successful physical replay are retained. For each sampled target pose $T_o=(p_o,R_o)$, we rank source demonstrations by spatial proximity:
\begin{equation}
d_i = w_p \lVert p_o-p_o^{\mathrm{src},i}\rVert_2 + w_R d_R(R_o,R_o^{\mathrm{src},i}),
\label{eq}
\end{equation}
where $d_R$ measures the relative rotation angle, and $w_p,w_R$ weight positional and rotational differences. During data generation, we randomly select one of the $K_{\mathrm{src}}$ nearest valid demonstrations and adapt its interaction segments to the sampled target.

\paragraph{Stage 1: Demonstration-Anchored Whole-Body Motion Synthesis.}
For each generation attempt, we sample a task configuration around the source setup, varying the robot interaction pose and relevant object or placement-target poses. The initial walking position is sampled from navigable space. We then synthesize a continuous whole-body reference by replanning locomotion and adapting source interaction motions to the sampled configuration.

\textit{For locomotion segments}, A* plans collision-free paths between the required robot poses. The SONIC motion planner converts these paths into whole-body reference motions using a walking speed sampled once per trajectory from $[0.5,0.7]$\,m/s and a four-frame motion context. For vertical adjustments, such as squatting, we use the planner to generate root-height transitions. When a sustained upper-body posture is required, such as carrying an object, we preserve the corresponding joint angles in both the planning context and generated reference. The resulting reference updates the motion context for subsequent planning. \textit{For leg-critical interactions}, we retain the demonstrated lower-body motion and combine it with the preceding segment's terminal upper-body posture.

\textit{For manipulation segments}, we adapt source hand trajectories to the sampled interaction targets while keeping the reference root pose fixed. Let $T_o^{\mathrm{src}},T_o\in SE(3)$ denote the source and sampled poses of an interaction object or placement target, and $T_{e,t}^{\mathrm{src}}\in SE(3)$ the source hand pose, all expressed in the world frame. The adapted hand target is
\begin{equation}
\widetilde T_{e,t}
=
T_o(T_o^{\mathrm{src}})^{-1}T_{e,t}^{\mathrm{src}}.
\label{eq:method_object_relative_retargeting}
\end{equation}
This transformation preserves the demonstrated hand--target relation under the new configuration. For placement, $T_o$ represents the receptacle target; when its orientation is unchanged, adaptation reduces to a positional shift while retaining the source hand orientation. Given a baseline reference pose $r_t^{\mathrm{base}}$ from the selected source or preceding segment, arm inverse kinematics (IK) updates the active arm joints:
\begin{equation}
\widetilde q_t^{\mathrm{arm}} = \operatorname{IK}_{\mathrm{arm}}(\widetilde T_{e,t};r_t^{\mathrm{base}}).
\label{eq:target_arm_reference}
\end{equation}
The root pose and inactive joints remain unchanged during reference construction, while their executed states are determined by the subsequent whole-body rollout.

To ensure continuity between segments, we align their boundary poses and adjust transition timing. Each segment inherits the final reference pose of its predecessor. We connect this inherited pose to the retargeted manipulation trajectory using a gradually decaying hand-position offset:
\begin{equation}
p_{e,t}^{\mathrm{blend}}
=
\widetilde p_{e,t}+
\lambda_t\bigl(p_e^{\mathrm{in}}-\widetilde p_{e,t_0}\bigr).
\label{eq:method_transition_blend}
\end{equation}
Here, $\widetilde p_{e,t}$ is the position component of $\widetilde T_{e,t}$, $p_e^{\mathrm{in}}$ is the inherited hand position, and $t_0$ marks the transition start. As $\lambda_t$ decreases smoothly from one to zero, the hand moves from its inherited position onto the adapted trajectory. Since retargeting may increase the displacement between segments, we adjust transition durations according to the source hand's linear and angular speeds, keeping grasp and release commands synchronized with the retimed motion. This timing adjustment is independent of the sampled walking speed. Initial hand perturbations are defined in the robot-local frame and incorporated through arm IK. After grasping, connecting motions lift the object before moving toward the next target to reduce contact with the supporting surface.

\paragraph{Stage 2: Physics-Grounded Rollout Validation.}
We execute each synthesized reference motion in simulation to evaluate its feasibility under the deployment controller. At each step, the SONIC encoder maps a reference-motion window $\mathcal{R}_t$ to a 64-dimensional latent command $z_t=E(\mathcal{R}_t)$. The decoder combines this command with proprioceptive history $h_t$ to produce joint targets $u_t=D(z_t,h_t)$ for low-level PD control. Two hand commands $g_t\in[0,1]^2$ are executed separately. We record the resulting actions $a_t=(z_t,g_t)\in\mathbb{R}^{66}$ together with the executed robot and object states.
Each rollout is evaluated against task success, stability, and collision criteria. Successful executions are added to $\mathcal{D}$, while failed rollouts are examined for recoverable tracking errors. Deviations in the realized root, hand, or object poses, such as locomotion-endpoint or grasp-height errors, provide feedback for Stage 3. Falls, invalid collisions, and failures beyond the correction scope instead trigger resampling in Stage 1.

\paragraph{Stage 3: Execution-Conditioned Correction.}
Even when reference motions are kinematically valid, tracking deviations may alter the robot--object relations required for successful interaction. We therefore use rollout feedback to jointly correct the environment configuration and reference motion. Specifically, we compensate for planar tracking errors by adjusting relevant scene assets or interaction targets, while vertical grasp errors are corrected through reference-motion adaptation.

\textit{Robot-Pose-Guided $SE(2)$ Scene Correction.}
Locomotion tracking errors can shift the executed robot pose away from its intended interaction configuration. We apply an $SE(2)$ transformation to relevant scene assets to restore their intended spatial relations with the executed robot. Let $(x,\psi)$ and $(x',\psi')$ denote the intended and executed planar root poses, where $x\in\mathbb{R}^2$ and $\psi$ denotes heading. For a scene asset at $(x_j,\psi_j)$ in the same world frame, we apply
\begin{equation}
x_j^+ = x' + R_2(\psi'-\psi)(x_j-x),
\qquad
\psi_j^+ = \psi_j+\psi'-\psi .
\label{eq:method_robot_scene_correction}
\end{equation}
Here, $R_2(\cdot)$ denotes planar rotation, and the superscript $+$ indicates the corrected asset pose. This transformation preserves the intended robot--asset relationship relative to the executed root pose.

\textit{Object $SE(2)$ Correction.}
Scene-level alignment cannot fully correct local tracking errors during grasping or placement. We therefore adjust the interaction object or placement target using residual error relative to a successful source rollout:
\begin{equation}
\Delta e_{xy}
=
e_{xy}^{\mathrm{cand}}-e_{xy}^{\mathrm{src}}.
\label{eq:method_source_relative_error}
\end{equation}
Here, $e_{xy}^{\mathrm{cand}}$ and $e_{xy}^{\mathrm{src}}$ are the planar tracking errors of the candidate and successful source rollouts, measured at the corresponding grasp or release event. Each error is defined as the executed-minus-reference position of the interacting hand or carried-object anchor in a common planar frame. Their difference isolates the additional tracking deviation introduced in the candidate execution. We compensate by translating the interaction object or placement target by $\Delta e_{xy}$ and adjusting its yaw according to task-specific rules. Robot-relative targets retain their source yaw offset from the executed robot heading, while fixed-orientation targets preserve their configured yaw. 

\textit{Z-axis Compensation.}
Planar adjustments cannot resolve vertical wrist--object misalignment during grasping. We compare the relative wrist heights at grasp closure in the candidate and successful source rollouts. The resulting discrepancy adjusts the reference wrist height to restore the source-relative grasp geometry. This correction is applied smoothly around grasp closure and converted to arm joint targets through IK, while keeping the object height fixed.

After correction, the updated environment and reference motion are re-executed in Stage 2. Only rollouts satisfying the same task success, stability, and collision criteria are accepted. These validated executions provide synchronized robot and object states, together with action labels, for subsequent rendering and policy learning.

\subsection{Visual Observations and Real-World Deployment}
\label{sec:observation_rendering}
\label{sec:policy_learning}
We generate visual observations from physically validated executions rather than planned reference motions. By replaying the recorded robot and object states, we obtain synchronized egocentric views. We then combine Gaussian background renderings with simulator-rendered robots and objects through mask-guided z-buffer compositing to produce the final egocentric images.

Each rendered image $I_t$ and the corresponding proprioceptive state $\rho_t$ form an observation $o_t=(I_t,\rho_t)$, paired with the task instruction $\ell$ and recorded action $a_t=(z_t,g_t)$. We train a visuomotor policy to predict $H$-step action chunks, $(\hat z_{t:t+H-1},\hat g_{t:t+H-1})=\pi_\theta(o_t,\ell)$. At deployment, predicted latents drive the SONIC decoder, while hand commands are executed separately through the same interface used during data generation. See Section~\ref{app:system_setup} for deployment details.

\vspace{-8pt}
\section{Experiments}
\label{sec:experiments}
\vspace{-4pt}
In this section, we aim to answer four questions through real-robot experiments and simulation evaluation:

\begin{itemize}[leftmargin=*, nolistsep, noitemsep]
    \item \textbf{Q1: Data Generation Quality and Efficiency.} (\cref{sec:generation_efficiency}).
    How efficiently does \method generate successful demonstrations and how does each key component affect data quality?

    \item \textbf{Q2: Direct Sim-to-Real Transfer and Generalization.}   (\cref{sec:sim2real}).
    Can simulation data alone support direct sim-to-real transfer, and how do the resulting policies compare with those trained on real demonstrations? How do they scale and generalize across real-world configurations?

    \item \textbf{Q3: Understanding Sim-to-Real Transfer.} (\cref{sec:cotrain}).
    Can simulation reflect real-world performance? What gaps remain in actions and observations? Does mixing real and simulation data improve learning? 
    \item \textbf{Q4: Zero-HLM-Bench.} (\cref{app:benchmark_details}).
    How do different policy families perform under shared evaluation protocols, and how does performance scale with simulation data?
\end{itemize}

\subsection{Experimental Setup}
\label{sec:exp_setup}

\label{app:system_setup}
\paragraph{System Setup.}
\looseness=-1
We use a Unitree G1 with 29 body degrees of freedom and two Inspire Robots RH56E2 hands, each controlled through a single scalar opening command. A head-mounted RealSense D455 provides egocentric RGB observations with automatic exposure disabled to reduce visual variation. The visuomotor policy takes RGB observations and a 34-dimensional robot state consisting of 29 body joint positions, three projected-gravity components, and two hand-opening values. It predicts 66-dimensional actions comprising a 64-dimensional SONIC motion latent and two hand-opening commands. We use real-time chunking (RTC)~\citep{black2025rtc} for asynchronous inference and execution at 50\,Hz. The frozen SONIC decoder converts motion latents into body joint targets, while hand commands are executed through a separate interface. The policy and SONIC run on an RTX 4090; simulation uses the same robot and hand configuration.

Real-world demonstrations are collected through PICO whole-body teleoperation with the SONIC controller~\citep{luo2026sonic}. Simulation data generation runs on eight RTX 4090 GPUs, and all policies are trained on eight A100 GPUs. \Cref{fig:app_system_setup} summarizes the deployment pipeline.

\par\addvspace{6pt}
\noindent\begin{minipage}{\linewidth}
\centering
    \includegraphics[width=0.9\linewidth]{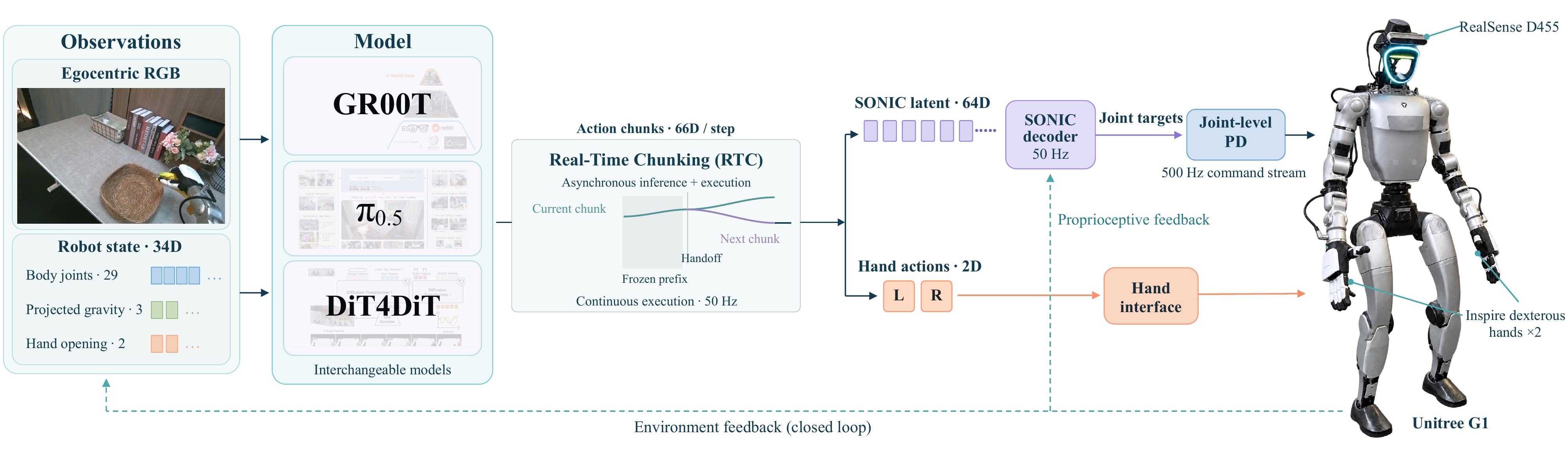}
    \captionof{figure}{\textbf{System Setup and Deployment.} The policy maps egocentric RGB observations and robot state to SONIC motion latents and hand-opening commands. The frozen SONIC decoder produces body joint targets, while the two hands are controlled through a separate interface.}
    \label{fig:app_system_setup}
\end{minipage}
\par\addvspace{6pt}

\paragraph{Tasks.}
We evaluate three long-horizon loco-manipulation tasks covering diverse whole-body skills:
\begin{itemize}[leftmargin=*,labelsep=0.5em]
    \item \textbf{Task A: Mobile Pick-and-Place.} The robot walks to the table, picks up a banana, and places it in a basket. The reported stages are Walk, Pick, and Place, where Place reports full-task success.
    \item \textbf{Task B: Low-Reach Hat Placement.} The robot squats to pick up a hat, stands, carries it to a cabinet, and places it on the cabinet. The reported stages are Squat-Pick, Carry, and Place, where Place reports full-task success.
    \item \textbf{Task C: Leg-Assisted Basket Transport.} The robot carries a fruit basket bimanually, closes an obstructing cabinet door with its foot, and places the basket on a stool. The reported stages are Bi-Pick, Kick, and Bi-Place, where Bi-Place reports full-task success.
\end{itemize}

Task sequences are illustrated in~\cref{fig:generalization_overview} (a).

\paragraph{Camera Calibration.}
We first define an upper-body motion trajectory that maintains visibility of the robot's arms and hands. The physical robot executes this trajectory while recording egocentric RGB video and joint configurations. From selected video frames, we annotate 2D arm and hand landmarks and associate them with corresponding link-local 3D points from the robot URDF. Forward kinematics transforms these points into the camera mounting-link frame using the recorded joint configurations $q_t$. Given fixed camera intrinsics $K$, we obtain per-frame extrinsic estimates through PnP and discard frames with large reprojection errors. We then jointly optimize a fixed transform $T$ from the mounting-link frame to the camera optical frame by minimizing the robust reprojection error:
\begin{equation}
    T^{\star}
    =\underset{T\in\mathrm{SE}(3)}{\operatorname{arg\,min}}
    \sum_{(t,i)\in\mathcal{V}}\sum_{k=1}^{2}
    \rho\!\left(
    \left[
    \pi_K\!\left(T\mathbf{x}_i(q_t)\right)-\mathbf{u}_{ti}
    \right]_k^2
    \right),
    \label{eq:camera_extrinsic_calibration}
\end{equation}
where $\mathbf{u}_{ti}$ denotes the annotated image location, $\pi_K$ is the camera projection function, and $\rho$ is a soft-$\ell_1$ loss.

\paragraph{Evaluation Configurations and Simulation Sampling.}
\label{app:evaluation_configurations}
To systematically evaluate spatial generalization, we define two types of task configurations:
\textsc{Base} denotes the reference configuration used for real-world demonstration collection, while \textsc{Expanded} introduces shifted robot starts and interaction locations for Tasks A and B; Task A additionally includes an alternative object--goal pair, whereas Task C uses only \textsc{Base}. As illustrated in \cref{fig:ood_position}, each \textsc{Base} or indexed \textsc{Expanded} location $[i,j]$ defines the center of a local sampling region. Robot starts are sampled within a $30\times30$\,cm region around each center, while picking and placement locations use a $5\times5$\,cm region. For simulation data generation, \cref{tab:object_sampling} summarizes the task-specific perturbations around these configurations. Notably, in Task B, the hat and cabinet are additionally perturbed by $\pm5$\,cm and $\pm40$\,cm per axis, respectively, with hat yaw varied by $\pm20^\circ$; the reference root height is lowered by 5--10\,cm relative to the source pose. In Task C, the robot and basket are perturbed around the source configuration, while the stool pose is coupled to the robot endpoint. All planar offsets are expressed in the world $XY$ frame; roll and pitch remain fixed unless otherwise specified.

\par\addvspace{6pt}
\noindent\begin{minipage}{\linewidth}
\centering
    \includegraphics[width=\linewidth,trim=0 30bp 0 0,clip]{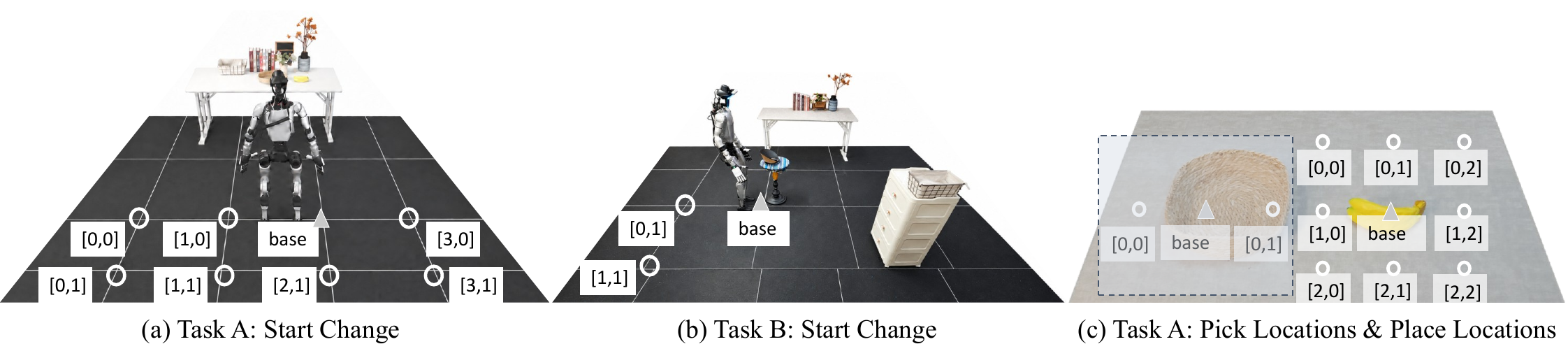}
    \vspace{2pt}
    \begin{minipage}[t]{0.345\linewidth}
        \centering\small
        (a) Robot-start sampling\\[2pt]
        \textsc{Base}: $30\times30$\,cm\\
        \textsc{Expanded} $[i,j]$: $30\times30$\,cm
    \end{minipage}\hfill
    \begin{minipage}[t]{0.315\linewidth}
        \centering\small
        (b) Robot-start sampling\\[2pt]
        \textsc{Base}: $30\times30$\,cm\\
        \textsc{Expanded} $[i,j]$: $30\times30$\,cm
    \end{minipage}\hfill
    \begin{minipage}[t]{0.325\linewidth}
        \centering\small
        (c) Picking / placement sampling\\[2pt]
        \textsc{Base}: $5\times5$\,cm\\
        \textsc{Expanded} $[i,j]$: $5\times5$\,cm
    \end{minipage}
    \captionsetup{skip=4pt}
    \captionof{figure}{\textbf{Base and Expanded Configurations.} (a,b) Robot starting locations for Tasks A and B. (c) Task A picking and placement locations. \textsc{Base} marks the reference configuration; \textsc{Expanded} $[i,j]$ identifies shifted configurations. The annotations below each panel give the full simulation sampling widths around its centers. Task A start indices follow [column, row].}
    \label{fig:ood_position}
\end{minipage}
\begin{table}[!htbp]
\centering
    \begin{minipage}{\linewidth}
    \centering
    \captionof{table}{\textbf{Simulation Configuration Sampling.} Offsets are relative to the corresponding \textsc{Base} or \textsc{Expanded} $[i,j]$ center. Endpoint widths describe a sampling region, rather than an independent object perturbation.}
    \label{tab:object_sampling}
    \small
    \setlength{\tabcolsep}{4pt}
    \renewcommand{\arraystretch}{1.08}
    \begin{tabularx}{\linewidth}{@{}lXccc@{}}
        \toprule
        Task & Entity / configuration & $\Delta X$ (cm) & $\Delta Y$ (cm) & $\Delta$Yaw ($^\circ$) \\
        \midrule
        A & Robot start: \textsc{Base} / \textsc{Expanded} $[i,j]$ (a) & $\pm15$ & $\pm15$ & $\pm15$ \\
          & Picking location: \textsc{Base} / \textsc{Expanded} $[i,j]$ (c) & $\pm2.5$ & $\pm2.5$ & $\pm15$ \\
          & Placement location: \textsc{Base} / \textsc{Expanded} $[i,j]$ (c) & $\pm2.5$ & $\pm2.5$ & $\pm20$ \\
          & Robot interaction pose & $\pm10$ & $\pm15$ & $\pm15$ \\
        \midrule
        B & Robot start: \textsc{Base} / \textsc{Expanded} $[i,j]$ (b) & $\pm15$ & $\pm15$ & $\pm15$ \\
          & Robot pickup pose & $\pm10$ & $\pm10$ & $\pm15$ \\
          & Pickup root-height offset & \multicolumn{3}{c@{}}{$\Delta z\in[-10,-5]$\,cm} \\
          & Hat (\textsc{Base}) & $\pm5$ & $\pm5$ & $\pm20$ \\
          & Cabinet (\textsc{Base}) & $\pm40$ & $\pm40$ & $0$ \\
        \midrule
        C & Robot interaction pose & $\pm15$ & $[-10,5]$ & $\pm15$ \\
          & Basket (\textsc{Base}) & $[-10,5]$ & $[0,10]$ & $0$ \\
          & Stool & \multicolumn{3}{c@{}}{Coupled to the robot endpoint} \\
        \bottomrule
    \end{tabularx}
    \par\vspace{3pt}
    \raggedright\footnotesize
    Zero denotes no additional yaw perturbation. For Task C, the robot endpoint is sampled over a $51.644\times39.66$\,cm region, and the stool preserves its source pose relative to that endpoint.
    \end{minipage}
\end{table}
\par\addvspace{6pt}

\paragraph{Data Settings and Evaluation Protocol.}
We compare policies trained on real-world demonstrations and simulation-generated data under \textsc{Base} and \textsc{Expanded} configurations, as illustrated in \cref{fig:generalization_overview}(b). The real-data baseline, Real-Only(100), uses 100 demonstrations per task, while all simulation-generation methods start from the same pool of 15 teleoperated source demonstrations per task. As an agentic baseline, GPT-6 Astra generates 100 demonstrations for Tasks A and B using source demonstrations and full simulator state, denoted Sim-Astra(100). Our method generates 800 \textsc{Base} demonstrations per task (Sim-Ours) and an additional 800 \textsc{Expanded} demonstrations for each of Tasks A and B (Sim-Ours$^{*}$). Together with 800 object--goal-shift demonstrations, we generate 4,800 successful demonstrations in total.

For policy learning, 
we fine-tune pretrained GR00T-N1.7~\citep{nvidia2025grootn1} under each data setting and deploy all policies through the same SONIC whole-body controller. Each policy is evaluated on the physical robot over 20 trials per task in \textsc{Base} and five trials per configuration in \textsc{Expanded}. We report stage-wise success rates for individual locomotion and interaction stages, alongside full-task success rates requiring completion of the entire task sequence. 

\begingroup
\setlength{\intextsep}{8pt plus 2pt minus 2pt}
\captionsetup{skip=4pt}
\setlength{\textfloatsep}{12pt plus 2pt minus 2pt}
\setlength{\floatsep}{8pt plus 2pt minus 2pt}
\begin{samepage}
\Needspace{9\baselineskip}
\subsection{Data Generation Quality and Efficiency}
\label{sec:generation_efficiency}
\label{app:ablation_studies}
To answer \textbf{Q1}, we evaluate reconstruction quality and cost (\cref{sec:reconstruction_cost}), demonstration generation success and efficiency (\cref{app:generation_diagnostics}), and the contributions of key components (\cref{app:rendering_ablation}).

\subsubsection{Reconstruction Cost and Geometric Accuracy}
\label{sec:reconstruction_cost}
\Cref{tab:agentic_resource_usage} reports the time and cost of scene construction, physical configuration, and seven interaction-asset reconstructions, totaling 7h 13m and \$22.82.
\end{samepage}
As shown in \cref{tab:reconstruction_geometry}, Gaussian guidance reduces the symmetric Chamfer distance to LiDAR ground truth from 11.689 to 5.517\,cm, outperforming both unguided reconstruction and GS-Mesh. Evaluation details and qualitative comparisons appear in Appendix~\ref{app:reconstruction_evaluation}.
\begin{table}[htbp]
\centering
\begin{minipage}[t]{0.59\linewidth}
\vspace{0pt}
\centering
\begin{minipage}[t][30pt][t]{\linewidth}
\captionof{table}{\textbf{Reconstruction Cost.} Time and allocated subscription cost by stage.}
\label{tab:agentic_resource_usage}
\end{minipage}\par
\footnotesize
\setlength{\tabcolsep}{6pt}
\renewcommand{\arraystretch}{1.12}
\begin{tabular*}{\linewidth}{@{\extracolsep{\fill}}lrr}
\toprule
Stage & Time & Cost (USD) \\
\midrule
Input $\rightarrow$ Blender & 1h 47m & 6.46 \\
Blender $\rightarrow$ IsaacLab & 1h 21m & 5.08 \\
Object reconstruction (7 assets) & 4h 05m & 11.28 \\
\bottomrule
\end{tabular*}
\end{minipage}\hfill
\begin{minipage}[t]{0.38\linewidth}
\vspace{0pt}
\centering
\begin{minipage}[t][30pt][t]{\linewidth}
\captionof{table}{\textbf{Geometry Accuracy.} Symmetric CD to LiDAR ground truth.}
\label{tab:reconstruction_geometry}
\end{minipage}\par
\footnotesize
\setlength{\tabcolsep}{6pt}
\renewcommand{\arraystretch}{1.12}
\begin{tabular*}{\linewidth}{@{\extracolsep{\fill}}lr}
\toprule
Method & CD (cm) $\downarrow$ \\
\midrule
GS-Mesh & 5.894 \\
Ours (w/o GS) & 11.689 \\
Ours & \textbf{5.517} \\
\bottomrule
\end{tabular*}
\end{minipage}
\end{table}

\subsubsection{Generation Success and Collection Time}
\label{app:generation_diagnostics}
We analyze the generation success and failure modes of Sim-Ours and Sim-Astra under the same task-success, stability, and collision criteria. As shown in \cref{fig:generation_failure_breakdown,fig:astra_generation_outcomes}, Sim-Ours achieves success rates of 33\%, 34\%, and 25\% on Tasks A--C, respectively, compared with 21.38\% and 24.70\% for Sim-Astra on Tasks A and B. Sim-Astra produces no successful Task C trajectories. Grasp-related failures dominate across tasks, while Sim-Astra additionally encounters invalid initialization, program errors, and interrupted runs. Each failed rollout is categorized by its earliest failed segment and a mutually exclusive failure type, with ribbon widths representing fractions of all attempts.

\begin{figure}[!htbp]
\centering
\begin{minipage}{0.96\linewidth}
\centering
\small
\includegraphics[width=\linewidth]{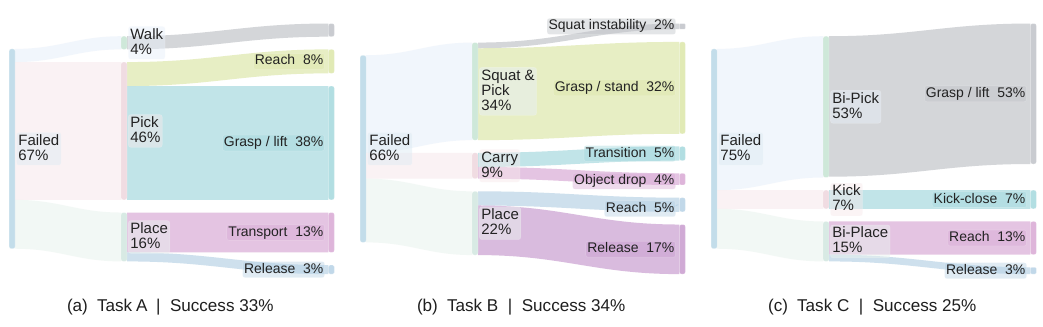}
\captionof{figure}{\textbf{Sim-Ours Generation Outcomes and Failure Breakdown.} Ribbons partition rollout attempts by task segment and outcome. Generation success is 33\%, 34\%, and 25\% for Tasks A, B, and C; ribbon percentages are rounded.}
\label{fig:generation_failure_breakdown}
\end{minipage}
\end{figure}
\begin{figure}[!htbp]
\centering
\begin{minipage}{0.96\linewidth}
\centering
\small
\includegraphics[width=\linewidth]{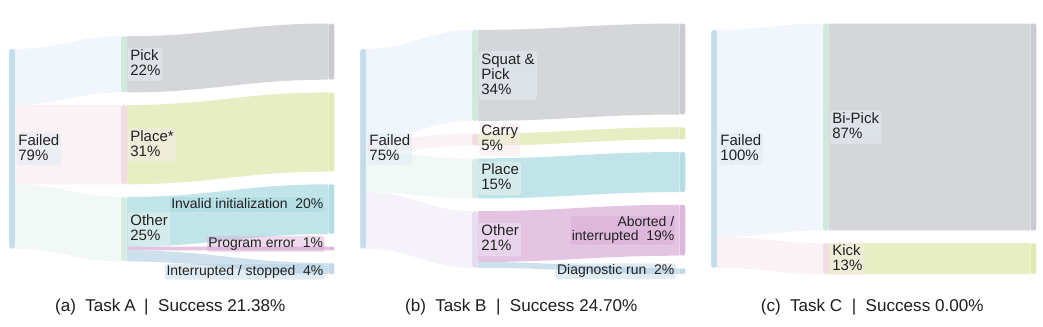}
\captionof{figure}{\textbf{GPT-6 Astra Generation Outcomes After Stabilization.} Success rates are 21.38\%, 24.70\%, and 0\% for Tasks A--C. Task C failures occur at Bi-Pick (87\%) and Kick (13\%); its total attempt count is unavailable. \textit{Place*} groups placement, hand-withdrawal, and acceptance failures.}
\label{fig:astra_generation_outcomes}
\end{minipage}
\end{figure}

We also compare the time required to collect 50 successful demonstrations (\cref{tab:sim_real_data_compare}). Sim-Ours achieves $2.4\times$, $1.6\times$, and $1.4\times$ speedups over real-world teleoperation on Tasks A--C, respectively, averaging $1.8\times$ across tasks. In contrast, Sim-Astra requires substantially longer collection times on Tasks A and B and produces no successful Task C demonstrations after more than eight hours.

\subsubsection{Data Generation Ablations}
\label{app:rendering_ablation}
\paragraph{Motion Adaptation Choices.}
We ablate motion adaptation and correction on Task A under \textsc{Base}, targeting 100 successful trajectories per variant. As shown in \cref{tab:ablation_generation_success}, removing robot-pose-guided $SE(2)$ correction causes the largest success drop, from 61\% to 16\%. Removing object-level $SE(2)$ correction, Z-axis compensation, or velocity-adaptive transitions reduces success to 49\%, 51\%, and 55\%, respectively. These results highlight the importance of scene correction and motion adaptation for reliable interaction.

\begin{table}[!htbp]
\centering
\begin{minipage}[t]{0.32\linewidth}
\vspace{0pt}
\centering
\begin{minipage}[t][32pt][t]{\linewidth}
\captionof{table}{\textbf{Collection Time.} Hours per 50 successes; Real excludes setup.}
\label{tab:sim_real_data_compare}
\end{minipage}\par
\footnotesize
\setlength{\tabcolsep}{2.5pt}
\renewcommand{\arraystretch}{1.12}
\begin{tabular*}{\linewidth}{@{\extracolsep{\fill}}lccc}
\toprule
Task & Real & \textbf{Sim-Ours} & Sim-Astra \\
\midrule
Task A & 1.35 & \textbf{0.56} & 5.00 \\[7.1pt]
Task B & 1.02 & \textbf{0.63} & 5.83 \\[7.1pt]
Task C & 1.68 & \textbf{1.19} & $>8.00$, fail \\[7.1pt]
\bottomrule
\end{tabular*}
\end{minipage}\hfill
\begin{minipage}[t]{0.37\linewidth}
\vspace{0pt}
\centering
\begin{minipage}[t][32pt][t]{\linewidth}
\captionof{table}{\textbf{Generation Ablations.} Task A, \textsc{Base}; 100 successes per variant.}
\label{tab:ablation_generation_success}
\end{minipage}\par
\footnotesize
\setlength{\tabcolsep}{2.5pt}
\renewcommand{\arraystretch}{1.12}
\begin{tabular*}{\linewidth}{@{\extracolsep{\fill}}lc}
\toprule
Variant & Success (\%) \\
\midrule
Full & \textbf{61.0} \\
w/o Velocity-Adaptive & 55.0 \\
w/o Z Compensation & 51.0 \\
w/o Object $SE(2)$ & 49.0 \\
w/o Robot $SE(2)$ & 16.0 \\
\bottomrule
\end{tabular*}
\end{minipage}
\hfill
\begin{minipage}[t]{0.28\linewidth}
\vspace{0pt}
\centering
\begin{minipage}[t][32pt][t]{\linewidth}
\captionof{table}{\textbf{Mesh vs. GS.} Task A real-world success.}
\label{tab:rendering_ablation}
\end{minipage}\par
\footnotesize
\setlength{\tabcolsep}{2.5pt}
\renewcommand{\arraystretch}{1.12}
\begin{tabular*}{\linewidth}{@{\extracolsep{\fill}}lcc}
\toprule
Stage & Mesh & GS \\
\midrule
Walk & 18/20 & \textbf{20/20} \\[7.1pt]
Pick & 3/20 & \textbf{10/20} \\[7.1pt]
Place & 3/20 & \textbf{7/20} \\[7.1pt]
\bottomrule
\end{tabular*}
\end{minipage}
\end{table}

\paragraph{Visual Observation Choices.}
With 100 simulated demonstrations per setting, we compare Gaussian and mesh rendering for real-world transfer. As shown in \cref{tab:rendering_ablation}, Gaussian observations improve Task A full-task success from 3/20 to 7/20. Picking improves substantially (3/20 to 10/20), while walking shows a smaller gain (18/20 to 20/20), suggesting that visual fidelity particularly benefits object interaction. \Cref{fig:rendering_comparison} compares both renderings with real-world observations.

\begingroup
\setlength{\intextsep}{4pt}
\begin{figure}[H]
\centering
\begin{minipage}{\linewidth}
\centering
\small
\includegraphics[width=0.94\linewidth]{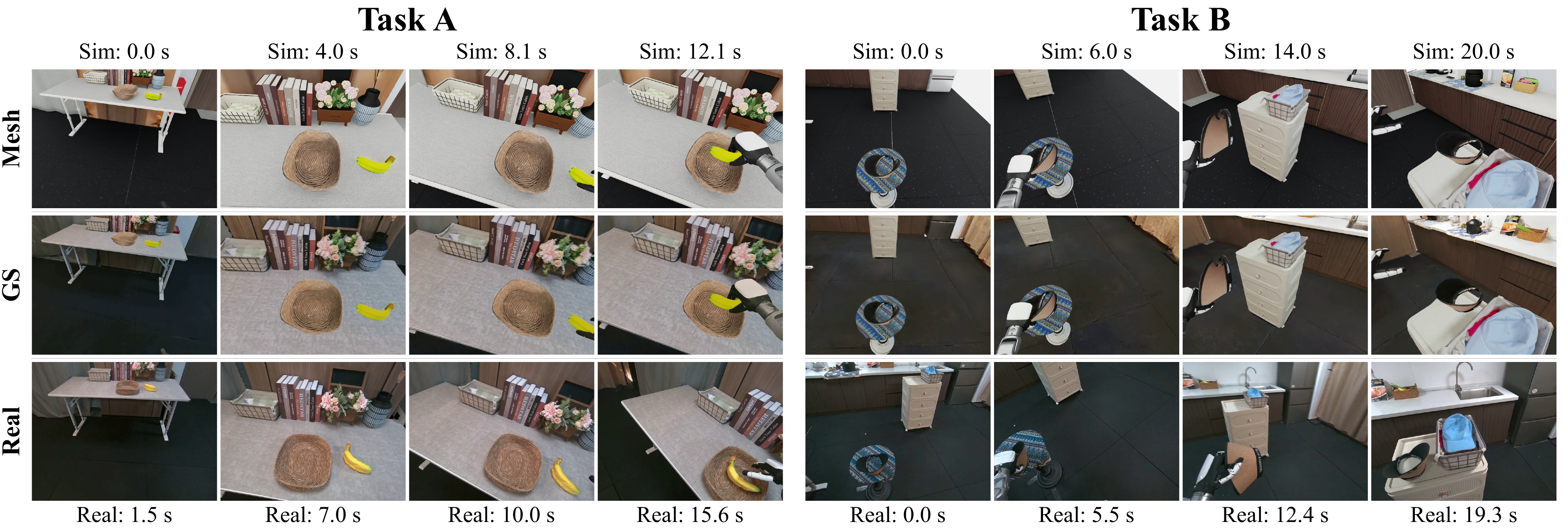}
\captionof{figure}{\textbf{Mesh, Gaussian-Splatting, and Real-World Observations.} Tasks A and B. Mesh and GS render the same simulation sequence; Real shows comparable stages from a separate physical rollout. Timestamps are independent across domains, not frame-aligned.}
\label{fig:rendering_comparison}
\end{minipage}
\end{figure}
\endgroup

\par\begingroup
\dimen0=\pagegoal\advance\dimen0 by -\pagetotal
\ifdim\dimen0<195pt\aftergroup\newpage\fi
\endgroup
\subsection{Direct Sim-to-Real Transfer and Generalization}
\label{sec:sim2real}

\paragraph{Direct Sim-to-Real Transfer.}
\begin{wrapfigure}{r}{0.43\linewidth}
\vspace{-8pt}
\centering
\includegraphics[width=\linewidth]{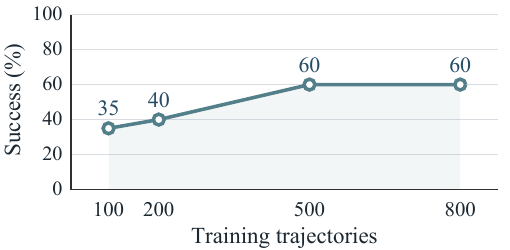}
\caption{\textbf{Sim-Data Scaling on Task A in Real-World Evaluation.}}
\label{fig:task_a_data_efficiency}
\vspace{-8pt}
\end{wrapfigure}
To answer \textbf{Q2}, we compare GR00T N1.7 fine-tuned on
Sim-Ours(100) and Sim-Ours(800) with Real-Only(100) and
Sim-Astra(100) under the \textsc{Base} configuration.
We evaluate all three tasks, except that Sim-Astra has
no usable Task C data.
As shown in~\cref{tab:sim2real_main}, Sim-Ours(100) achieves
mean full-task success comparable to Real-Only(100)
(43.3\% vs.\ 46.7\%), suggesting that our method provides
training supervision comparable to real-world demonstrations.
Increasing simulation data from 100 to 800 demonstrations per task raises
mean success across three tasks from 43.3\% to 66.7\%, surpassing Real-Only(100) at 46.7\%.
\Cref{fig:task_a_data_efficiency} further details how real-world
performance scales with the amount of our generated data on Task A.

\begin{table}[htbp]
\centering
\begin{minipage}{\linewidth}
\centering
\captionsetup{position=top, belowskip=0pt,skip=2pt}
\captionof{table}{%
\textbf{Direct Sim-to-Real Transfer under \textsc{Base}.}
We report stage-wise success, with locomotion stage in \textcolor{uclablue}{blue} and manipulation stage in \textcolor{taskred}{red}. Place and Bi-Place report full-task success. Avg. averages full-task success over Tasks A--C, except Sim-Astra (A--B only; Task C unavailable).}
\label{tab:sim2real_main}

\begingroup
\renewcommand{\arraystretch}{1.08}
\footnotesize
\compacttable

\begin{adjustbox}{max width=\linewidth}
\begin{tabular*}{\linewidth}{@{\extracolsep{\fill}}l ccc ccc ccc c@{}}
\toprule
\multirow{2}{*}{Training Data}
& \multicolumn{3}{c}{Task A}
& \multicolumn{3}{c}{Task B}
& \multicolumn{3}{c}{Task C}
& \multirow{2}{*}{\shortstack{Avg.}} \\
\cmidrule(lr){2-4}
\cmidrule(lr){5-7}
\cmidrule(lr){8-10}

& \textcolor{uclablue}{Walk}
& \textcolor{taskred}{Pick}
& \textcolor{taskred}{Place}
& \textcolor{taskred}{Sq-Pick}
& \textcolor{uclablue}{Carry}
& \textcolor{taskred}{Place}
& \textcolor{taskred}{Bi-Pick}
& \textcolor{uclablue}{Kick}
& \textcolor{taskred}{Bi-Place}
& \\

\midrule
Real-Only(100) & \textbf{20/20} & 10/20 & 8/20 & 16/20 & 16/20 & 11/20 & 15/20 & 13/20 & 9/20 & 46.7 \\
\addlinespace[1pt]
Sim-Astra(100) & 17/20 & 4/20 & 4/20 & 5/20 & 0/20 & 0/20 & --- & --- & --- & 10.0 \\
\addlinespace[1pt]
\textbf{Sim-Ours(100)} & \textbf{20/20} & 10/20 & 7/20 & 17/20 & 16/20 & 10/20 & 16/20 & 11/20 & 9/20 & 43.3 \\
\addlinespace[1pt]
\textbf{Sim-Ours(800)} & \textbf{20/20} & \textbf{15/20} & \textbf{12/20} & \textbf{19/20} & \textbf{19/20} & \textbf{15/20} & \textbf{20/20} & \textbf{19/20} & \textbf{13/20} & \textbf{66.7} \\

\bottomrule
\end{tabular*}
\end{adjustbox}

\endgroup
\end{minipage}
\par

\end{table}

\begingroup
\setlength{\intextsep}{4pt}
\begin{figure}[H]
\centering
\begin{minipage}{\linewidth}
\centering
\small
\includegraphics[width=\linewidth]{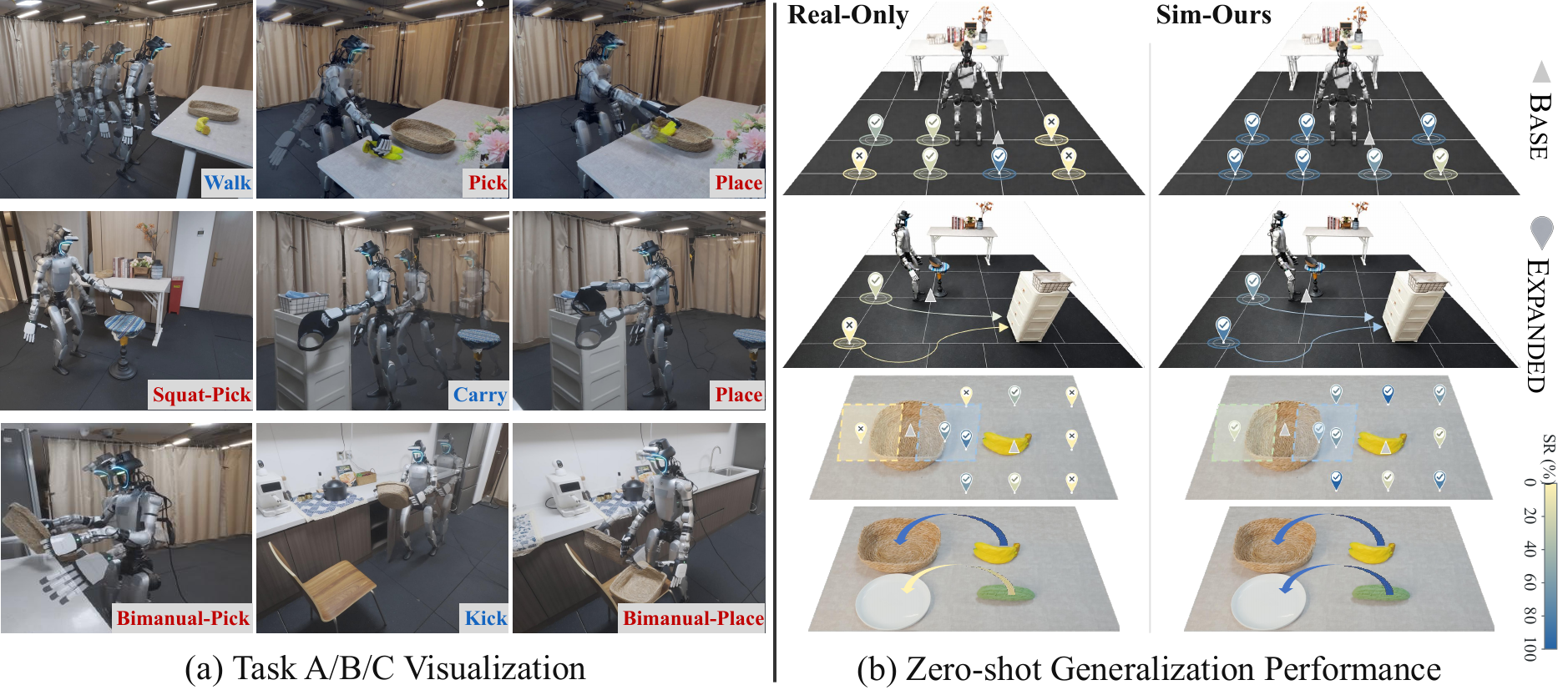}
\captionof{figure}{\textbf{Tasks and Generalization Evaluation.} (a) Task A--C stage sequences. (b) Robot starts for Tasks A and B, picking and placement locations, and an alternative object--goal pair (top to bottom). Triangles/pins mark \textsc{Base}/\textsc{Expanded}; colors show real-world successes per five trials, from yellow (0/5) to blue (5/5).}
\label{fig:generalization_overview}
\end{minipage}
\end{figure}
\endgroup

\paragraph{Generalization across Configurations.}

To assess whether expanded simulation coverage transfers to
real-world configurations beyond those covered by real demonstrations,
we compare Sim-Ours(800)$^{*}$ with Real-Only(100).
As shown in~\cref{tab:task_a_b_full_spatial},
Sim-Ours(800)$^{*}$ consistently outperforms Real-Only(100) across
all five \textsc{Expanded} settings, increasing pooled success from
23\% to 69\%.
The gains span robot-start shifts, picking and placement locations,
and the alternative object--goal pair, demonstrating that simulation
coverage transfers robustly across diverse configuration changes.
Notably, success on the object--goal shift increases from 20\% to
100\%.
These results show that scalable simulation data can substantially
broaden real-world policy coverage without requiring additional
physical demonstrations.
Detailed per-configuration results are provided in
Appendix~\ref{app:generalization}.

\par\addvspace{8pt}\noindent
\begin{minipage}[t]{0.64\linewidth}
\vspace{0pt}\centering
\begin{minipage}[t][42pt][t]{\linewidth}
\captionof{table}{\textbf{Zero-Shot Transfer across Configurations.} Real-world success rates (\%) across robot-start, tabletop-location, and object--goal shifts, excluding Base. Avg. pools successful trials across all five settings (100 total trials).}
\label{tab:task_a_b_full_spatial}
\end{minipage}\par
\small\renewcommand{\arraystretch}{1.45}\setlength{\tabcolsep}{2pt}\begin{tabular*}{\linewidth}{@{\extracolsep{\fill}}lcccccc@{}}
    \toprule
    \multirow{2}{*}{Training Data}
    & \multicolumn{2}{c}{Robot Start}
    & \multicolumn{2}{c}{Tabletop Location}
    & \multirow{2}{*}{\hspace{5pt}Object--Goal} & \multirow{2}{*}{Avg.} \\
    \cmidrule(lr){2-3}
    \cmidrule(lr){4-5}
      & Task A & Task B & Pick & Place & & \\
    \midrule
      Real-Only(100) & 22.9 & 10.0 & 25.0 & 30.0 & 20.0 & 23.0 \\
      \textbf{Sim-Ours(800)}$^{*}$ & \textbf{71.4} & \textbf{70.0} & \textbf{67.5} & \textbf{50.0} & \textbf{100.0} & \textbf{69.0} \\
    \bottomrule
    \end{tabular*}
\end{minipage}\hfill
\begin{minipage}[t]{0.33\linewidth}
\vspace{0pt}\centering
\begin{minipage}[t][42pt][t]{\linewidth}
\captionof{table}{\textbf{Sim--Real Consistency on Task A.} Full-task successes under matched \textsc{Base}; denominators give trial counts.}
\label{tab:benchmark_validate}
\end{minipage}\par
\small
\setlength{\tabcolsep}{3pt}
\renewcommand{\arraystretch}{1.20}
\begin{tabular*}{\linewidth}{@{\extracolsep{\fill}}llc@{}}
\toprule
Training Data & Test Domain & Success \\
\midrule
\multirow{2}{*}{Real(100)} & Real & 8/20 \\
 & Sim & 31/100 \\
\midrule
\multirow{2}{*}{Sim(100)} & Real & 7/20 \\
 & Sim & 39/100 \\
\bottomrule
\end{tabular*}
\end{minipage}
\par\addvspace{6pt}
\FloatBarrier
\subsection{Understanding Sim-to-Real Transfer}
\label{sec:cotrain}

\paragraph{Can Simulation Reflect Real-World Performance?}
We examine sim--real consistency using Zero-HLM-Bench (Section~\ref{app:benchmark_details}). Specifically, we evaluate Task A policies trained on Real-Only(100) and Sim-Ours(100) in both simulation and the real world under matched \textsc{Base} configurations. As shown in \cref{tab:benchmark_validate}, Real-Only(100) achieves 40\% real-world success versus 31\% in simulation, while Sim-Ours(100) achieves 35\% and 39\%, respectively. The corresponding sim--real gaps are 9 and 4 percentage points.

These results suggest that the benchmark captures a comparable performance range across both training sources without consistently overestimating real-world success. However, the reversed policy ranking indicates that agreement in absolute success rates does not guarantee consistent relative performance. Thus, Zero-HLM-Bench provides an approximate measure of task difficulty under matched configurations, rather than an exact predictor of policy rankings. Further benchmark evaluations, including model comparisons and data scaling, are presented in Section~\ref{app:benchmark_details}.

\paragraph{What Gaps Remain in Actions and Observations?}

To examine the transferability of simulated actions and visual observations, we evaluate the banana pick-and-place subtask in Task A, varying its picking location to construct four training sets (\cref{fig:data_compositions}).
 R-25 contains 25 real demonstrations with the banana placed in the lower region of the table, while S-25 contains 25 simulation demonstrations with the banana in the upper region. Mix-50 combines both datasets to cover both picking regions. For Mix-50*, we successfully replay the S-25 trajectories on the physical robot by manually adjusting object placements to match the simulated interactions. We then replace the simulated observations in S-25 with recorded real images while retaining the original simulation action labels. Combining these trajectories with R-25 yields Mix-50*, which preserves the same action supervision as Mix-50.

As shown in \cref{tab:data_compositions}, R-25 and S-25 fail to generalize to their respective unseen picking regions, achieving 0/5 success. Although Mix-50 covers both regions, it achieves only 2/5 and 4/5 success in the upper and lower regions, respectively. Notably, upper-region performance drops from 4/5 for S-25 to 2/5 after mixing, suggesting that broader spatial coverage alone does not ensure consistent improvements. In contrast, Mix-50* achieves 5/5 success in both regions without changing the demonstration count or action supervision. These results support the transferability of simulation-generated action labels while identifying visual mismatch as an important limitation in sim--real co-training.

\begin{figure}[H]
\centering
\begin{minipage}[t]{0.43\linewidth}
\vspace{0pt}
\centering
\captionsetup{position=top}
\captionof{figure}{\textbf{Spatial Coverage of Data.} S-25 covers the upper table region; R-25 covers the lower region. Mix combines both.}
\label{fig:data_compositions}
\includegraphics[width=0.97\linewidth]{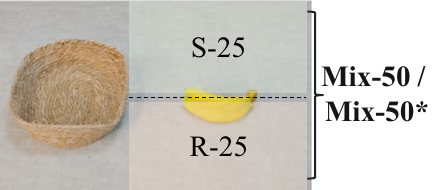}
\end{minipage}\hfill
\begin{minipage}[t]{0.53\linewidth}
\vspace{0pt}
\centering
\captionof{table}{\textbf{Different Data Compositions.} R denotes real-world data; S denotes simulation data.}
\label{tab:data_compositions}
\normalsize
\setlength{\tabcolsep}{5pt}
\renewcommand{\arraystretch}{1.48}
\begin{tabular*}{\linewidth}{@{\extracolsep{\fill}}lcccc@{}}
\toprule
Data & Vision & Action & $\mathrm{SR}_{\mathrm{Up}}$ & $\mathrm{SR}_{\mathrm{Low}}$ \\
\midrule
R-25 & R & R & 0/5 & 2/5 \\
S-25 & S & S & 4/5 & 0/5 \\
Mix-50 & S+R & S+R & 2/5 & 4/5 \\
Mix-50* & R & S+R & \textbf{5/5} & \textbf{5/5} \\
\bottomrule
\end{tabular*}
\end{minipage}
\end{figure}
\clearpage
\paragraph{Does Mixing Real and Simulation Data Improve Learning?}
\label{app:cotrain_details}

\begin{wrapfigure}{r}{0.43\linewidth}
\vspace{-8pt}
\centering
\includegraphics[width=\linewidth]{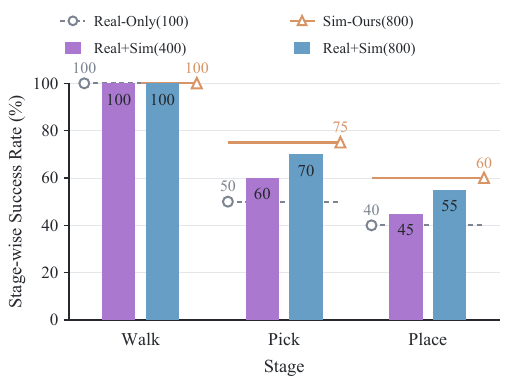}
\caption{\textbf{Co-Training Performance on Task A.}}
\label{fig:cotrain_performance}
\vspace{-8pt}
\end{wrapfigure}

We next study sim--real co-training under the \textsc{Base}
configuration, where real and simulation data follow matched
spatial and trajectory distributions.
Starting from 100 real demonstrations, we add either 400 or 800
simulation trajectories, with 20 real-world evaluation trials per task.

As shown in~\cref{fig:cotrain_performance}, increasing the amount
of simulation data improves Task A full-task success from 45\% to
55\%, but both co-training settings remain below Sim-Ours(800).
\Cref{tab:cotrain_stagewise} reports the full stage-wise results for Tasks A--C. Increasing simulation data improves full-task success from 9/20 to 11/20 on Task A, 11/20 to 12/20 on Task B, and 8/20 to 9/20 on Task C. Intermediate-stage performance does not improve uniformly: Task B Squat-Pick and Carry and Task C Kick decrease slightly.
Thus, simulation data benefits co-training, yet combining real and
simulation data does not necessarily outperform simulation-only
training.
Together with \cref{tab:data_compositions}, these results suggest
that the remaining visual-domain gap may limit effective sim--real
co-training.

\par\addvspace{6pt}
\noindent\begin{minipage}{\linewidth}
    \centering
    \captionof{table}{\textbf{Detailed Real--Sim Co-Training Results.} Entries are successful trials out of 20. Avg.\ is the mean full-task success rate across Tasks A--C. Both settings use 100 real demonstrations per task; parentheses give the added simulation count.}
    \label{tab:cotrain_stagewise}
    \footnotesize
    \setlength{\tabcolsep}{2pt}
    \renewcommand{\arraystretch}{1.1}
    \begin{adjustbox}{max width=\linewidth}
    \begin{tabular*}{\linewidth}{@{\extracolsep{\fill}}lc*{9}{c}c}
        \toprule
        Training Data & Real:Sim
        & \multicolumn{3}{c}{Task A}
        & \multicolumn{3}{c}{Task B}
        & \multicolumn{3}{c}{Task C}
        & Avg.\ (\%) \\
        \cmidrule(lr){3-5}\cmidrule(lr){6-8}\cmidrule(lr){9-11}
        & & \textcolor{uclablue}{Walk} & \textcolor{taskred}{Pick} & \textcolor{taskred}{Place}
        & \textcolor{taskred}{Sq-Pick} & \textcolor{uclablue}{Carry} & \textcolor{taskred}{Place}
        & \textcolor{taskred}{Bi-Pick} & \textcolor{uclablue}{Kick} & \textcolor{taskred}{Bi-Place} & \\
        \midrule
        Real+Sim(400) & 1:4
        & \textbf{20/20} & 12/20 & 9/20
        & \textbf{15/20} & \textbf{15/20} & 11/20
        & 18/20 & \textbf{13/20} & 8/20 & 46.7 \\
        Real+Sim(800) & 1:8
        & \textbf{20/20} & \textbf{14/20} & \textbf{11/20}
        & 14/20 & 14/20 & \textbf{12/20}
        & \textbf{20/20} & 12/20 & \textbf{9/20} & \textbf{53.3} \\
        \bottomrule
    \end{tabular*}
    \end{adjustbox}
\end{minipage}
\par\addvspace{6pt}

\FloatBarrier
\endgroup

\subsection{Zero-HLM-Bench: Humanoid Loco-Manipulation Evaluation}
\label{app:benchmark_details}

We introduce \textbf{Zero-HLM-Bench}, a humanoid loco-manipulation benchmark comprising the three task environments and nine skill stages under shared execution and evaluation protocols. We compare task-specific imitation learning, pretrained vision--language--action policies, and video--action modeling using the same generated training data.

\subsubsection{Simulation Setup and Shared Interface}
\label{app:benchmark_simulation_setup}

The benchmark instantiates the three task environments in IsaacLab with a simulated Unitree G1. Each policy receives egocentric RGB observations at $640\times480$ resolution and a 34-dimensional robot state consisting of 29 body joint positions, three projected-gravity components, and two hand states. Language-conditioned models additionally receive a fixed task instruction. Privileged object poses and contact states are used only for evaluation, never as policy inputs.

All models predict the same 66-dimensional action representation, consisting of a 64-dimensional SONIC motion latent and two hand commands. A frozen SONIC decoder converts the motion latent into body joint targets. Closed-loop rollouts run at 50\,Hz for at most 2,000 control steps (40\,s), with early termination upon task success or robot fall.

\begin{figure}[!htbp]
\begin{minipage}{\linewidth}
\centering
    \includegraphics[width=\linewidth]{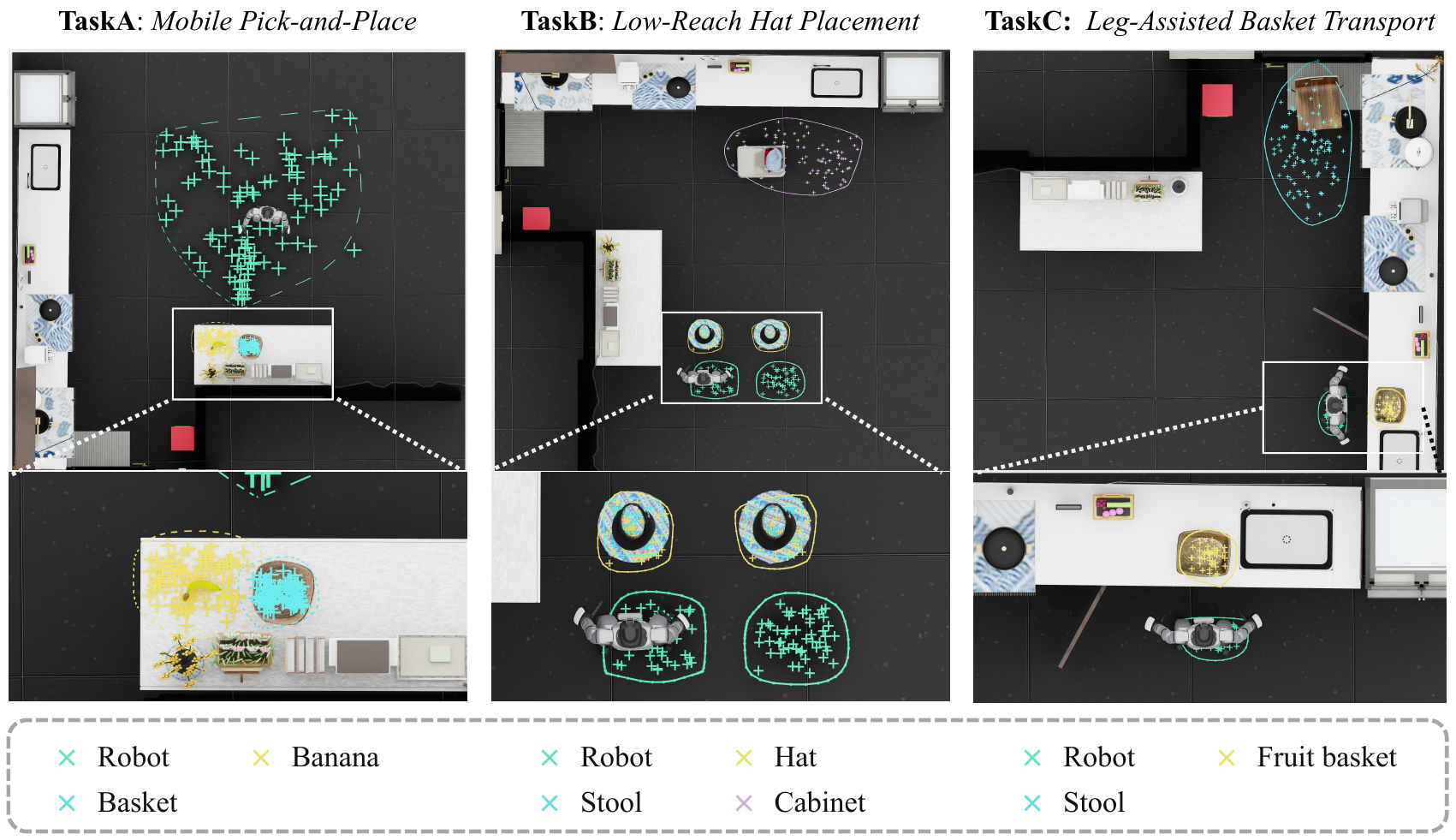}
    \captionof{figure}{\textbf{Benchmark Setup.} Top-down layouts and sampled robot and object configurations for Tasks A--C, with enlarged views of the interaction regions.}
    \label{fig:app_benchmark_setup}
\end{minipage}
\end{figure}

All models are evaluated using the same task-specific initialization distributions. \Cref{fig:app_benchmark_setup} visualizes the task layouts and sampled configurations used for benchmark evaluation.

\subsubsection{Model Training and Execution}
\label{app:benchmark_model_training}

All model comparisons use Sim-Ours(800)$^{*}$ for Tasks A and B, and Sim-Ours(800) for Task C, with 800 generated trajectories per task.
We retain each model's native training recipe while adapting its observation and action interfaces to the shared benchmark format. Image preprocessing, normalization, and padding follow the corresponding pretrained checkpoint.

\paragraph{Diffusion Policy.}
We adopt the LeRobot implementation of Diffusion Policy~\citep{chi2023diffusionpolicy} for task-specific imitation learning using RGB observations, robot states, and action sequences. Images are resized to $224\times224$, with robot states and actions normalized during training. The policy executes 16 actions per update.

\paragraph{$\pi_{0.5}$.}
We fine-tune the pretrained OpenPI $\pi_{0.5}$ checkpoint~\citep{black2025pi05} using RGB observations, robot states, and a fixed task instruction. The model predicts a 50-step horizon of 66-dimensional actions.

\paragraph{GR00T N1.7.}
We adapt pretrained GR00T N1.7~\citep{nvidia2025grootn1} to the G1--SONIC embodiment. Training updates the action-diffusion module, projection layers, and selected vision--language normalization layers, while keeping both backbones frozen. The model predicts a 40-step action horizon.

\paragraph{DiT4DiT.}
DiT4DiT~\citep{ma2026dit4dit} combines a pretrained Cosmos video backbone with an action-diffusion transformer. We follow its native video--action training recipe with a 40-step horizon of 66-dimensional actions.

We distinguish prediction horizons from execution chunks. Diffusion Policy executes 16 actions per update across all tasks, while the VLA and video--action models execute 16 on Task A and 40 on Tasks B and C.

\subsubsection{Evaluation Protocol and Metrics}
\label{app:benchmark_evaluation_metrics}

Each model--task pair is evaluated over 100 simulation trials. We report full-task success rates alongside independently evaluated stage-wise performance to assess individual skill capabilities. Unlike full-task success, stage-wise evaluations measure each skill independently rather than requiring successful completion of all preceding stages.

\paragraph{Task A.}
Walk requires the robot to reach the designated target region without a fall or execution error. Pick requires lifting the object by at least 3\,cm. Full-task success additionally requires the object to be placed inside the basket target region while satisfying the configured height and final-speed conditions.

\paragraph{Task B.}
Squat-Pick requires lifting the hat by at least 10\,cm. Carry requires the hat to enter the target region above the configured minimum height. Full-task success requires the hat to remain within the cabinet-local placement region with sufficiently low linear and angular velocity over the stability window.

\paragraph{Task C.}
Bi-Pick requires lifting the basket by at least 10\,cm. Kick records successful activation of the cabinet-door condition. Bi-Place reports full-task success and requires the door condition to hold while the basket is supported and settled within the stool-seat region over the stability window.

The reported Avg.\ is computed as the arithmetic mean of full-task success rates across Tasks A--C, excluding intermediate-stage evaluations.

\paragraph{Sim--Real Consistency.}
For the separate comparison in \cref{tab:benchmark_validate}, we evaluate the same Task A policies in simulation and the real world under the \textsc{Base} configuration. Each policy is tested in 100 simulation and 20 real-world trials.

\subsubsection{Stage-Wise Benchmark Results}
\label{app:benchmark_stagewise_results}

\Cref{tab:app_policy_benchmark_stagewise} reports the complete stage-wise benchmark results. Intermediate columns provide diagnostic success rates for individual task stages, while Place for Tasks A and B and Bi-Place for Task C denote full-task success.

\par\addvspace{6pt}
\noindent\begin{minipage}{\linewidth}
\centering
    \captionof{table}{\textbf{Stage-Wise Simulation Benchmark.} Stage-wise successes are reported as counts out of 100 trials per model--task pair. Intermediate stages are evaluated independently. Place for Tasks A and B and Bi-Place for Task C denote full-task success; Avg.\ is the mean full-task success rate (\%) across the three tasks.}
    \label{tab:app_policy_benchmark_stagewise}
    \label{tab:policy-benchmark}
    \footnotesize
    \compacttable
    \renewcommand{\arraystretch}{1.08}
    \begin{adjustbox}{max width=\linewidth}
    \begin{tabular*}{\linewidth}{@{\extracolsep{\fill}}l ccc ccc ccc c@{}}
        \toprule
        \multirow{2}{*}{Model}
        & \multicolumn{3}{c}{Task A}
        & \multicolumn{3}{c}{Task B}
        & \multicolumn{3}{c}{Task C}
        & \multirow{2}{*}{Avg.} \\
        \cmidrule(lr){2-4}\cmidrule(lr){5-7}\cmidrule(lr){8-10}
        & \textcolor{uclablue}{Walk} & \textcolor{taskred}{Pick} & \textcolor{taskred}{Place}
        & \textcolor{taskred}{Sq-Pick} & \textcolor{uclablue}{Carry} & \textcolor{taskred}{Place}
        & \textcolor{taskred}{Bi-Pick} & \textcolor{uclablue}{Kick} & \textcolor{taskred}{Bi-Place} & \\
        \midrule
        Diffusion Policy & 90/100 & 11/100 & 3/100 & 32/100 & 4/100 & 4/100 & 36/100 & 44/100 & 1/100 & 2.7 \\
        GR00T N1.7 & 100/100 & 85/100 & 66/100 & 92/100 & 76/100 & 74/100 & 93/100 & 97/100 & 69/100 & 69.7 \\
        $\pi_{0.5}$ & 100/100 & 82/100 & 60/100 & 79/100 & 60/100 & 60/100 & 93/100 & 98/100 & 57/100 & 59.0 \\
        DiT4DiT & 97/100 & 76/100 & 58/100 & 100/100 & 84/100 & 80/100 & 97/100 & 99/100 & 54/100 & 64.0 \\
        \bottomrule
    \end{tabular*}
    \end{adjustbox}
\end{minipage}
\par\addvspace{6pt}
\noindent\begin{minipage}[t]{0.52\linewidth}
\vspace{0pt}
\setlength{\baselineskip}{12.4pt}
\textbf{Policy Comparison and Data Scaling.} 
\Cref{tab:policy-benchmark} compares four policies under the same evaluation protocol. GR00T N1.7 achieves the highest mean full-task success (69.7\%), while DiT4DiT performs best on Task B (80\%). \Cref{fig:benchmark_scaling} shows consistent gains with increasing simulation data for GR00T N1.7. The gains extend across all three tasks, indicating that additional generated demonstrations benefit distinct whole-body interaction sequences. Improvements are larger at smaller data scales and become more gradual at higher counts; thus, the benefit is shared across tasks but is not linear in dataset size. The policy comparison also shows that high success on an individual stage does not guarantee high full-task success, motivating evaluation of both stage-wise behavior and complete execution. For example, Diffusion Policy reaches 90\% walking success on Task A but only 3\% full-task success.
\end{minipage}\hfill
\begin{minipage}[t]{0.44\linewidth}
\vspace{0pt}
\centering
\includegraphics[width=\linewidth]{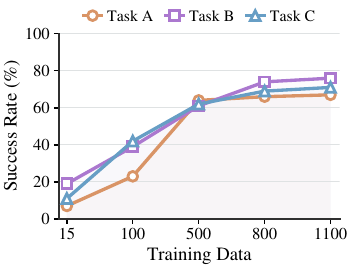}
\captionof{figure}{\textbf{Sim-Data Scaling in the Benchmark.} Full-task success improves with more generated demonstrations.}
\label{fig:benchmark_scaling}
\end{minipage}

\section{Conclusions}
\label{sec:conclusions}

We presented \method, a simulation-based data-generation framework that provides a practical path toward generalizable real-world humanoid loco-manipulation. Across three long-horizon tasks, policies trained solely on our simulation data transfer zero-shot to the real world and achieve higher mean full-task success than their real-data-trained counterparts. Expanding simulation coverage further improves transfer across spatial configurations without additional real demonstrations, showing that scalable synthetic coverage can translate into broader real-world policy generalization. Our transfer analysis supports the effectiveness of generated action labels while revealing a remaining visual-domain gap. The co-training results show that mixing real and simulation data does not necessarily outperform simulation-only training, underscoring the importance of visual alignment for effective co-training. Future work will extend the framework to additional unified whole-body controllers and examine how tracking accuracy and motion capability affect demonstration quality. Another important direction is to extend the benchmark to richer interaction skills and multiple reconstructed scenes, enabling a more systematic study of cross-scene policy generalization.

\bibliographystyle{\paperbibstyle}
\bibliography{references}
\clearpage

\appendix
\begingroup
\raggedbottom
\setlength{\intextsep}{6pt plus 1pt minus 1pt}
\captionsetup{skip=4pt,belowskip=0pt}
\section*{Appendix Contents}
\begin{itemize}[leftmargin=*,labelsep=0.5em,itemsep=2pt,parsep=0pt]
    \item \textbf{\hyperref[app:method_details]{\ref*{app:method_details} Method Implementation Details}}
    \begin{itemize}[leftmargin=*,nosep]
        \item \hyperref[app:agentic_real2sim]{\ref*{app:agentic_real2sim} Agentic Reconstruction: Implementation Details}
    \end{itemize}
    \item \textbf{\hyperref[app:experimental_details]{\ref*{app:experimental_details} Experimental Details}}
    \begin{itemize}[leftmargin=*,nosep]
        \item \hyperref[app:reconstruction_evaluation]{\ref*{app:reconstruction_evaluation} Agentic Reconstruction Evaluation}
        \item \hyperref[app:astra_baseline]{\ref*{app:astra_baseline} GPT-6 Astra Baseline}
        \item \hyperref[app:generalization]{\ref*{app:generalization} Generalization Evaluation Details}
    \end{itemize}
    \item \textbf{\hyperref[app:qualitative_results]{\ref*{app:qualitative_results} Additional Qualitative Results}}
\end{itemize}

\section{Method Implementation Details}
\label{app:method_details}
We provide implementation details that complement the method
described in \cref{sec:method}.
Section~\ref{sec:trajectory_generation} details source annotation, source selection, and reference-motion construction; this appendix illustrates reconstruction.

\subsection{Agentic Reconstruction: Implementation Details}
\label{app:agentic_real2sim}

This section provides implementation details for the reconstruction pipeline in \cref{sec:real2sim}. \Cref{fig:agentic_real2sim_comparison} illustrates the workflow from captured inputs to a physics-ready simulation environment. See Section~\ref{sec:reconstruction_cost} for reconstruction time and cost.

\par\addvspace{6pt}
\noindent\begin{minipage}{\linewidth}
\centering
    \includegraphics[width=0.92\linewidth]{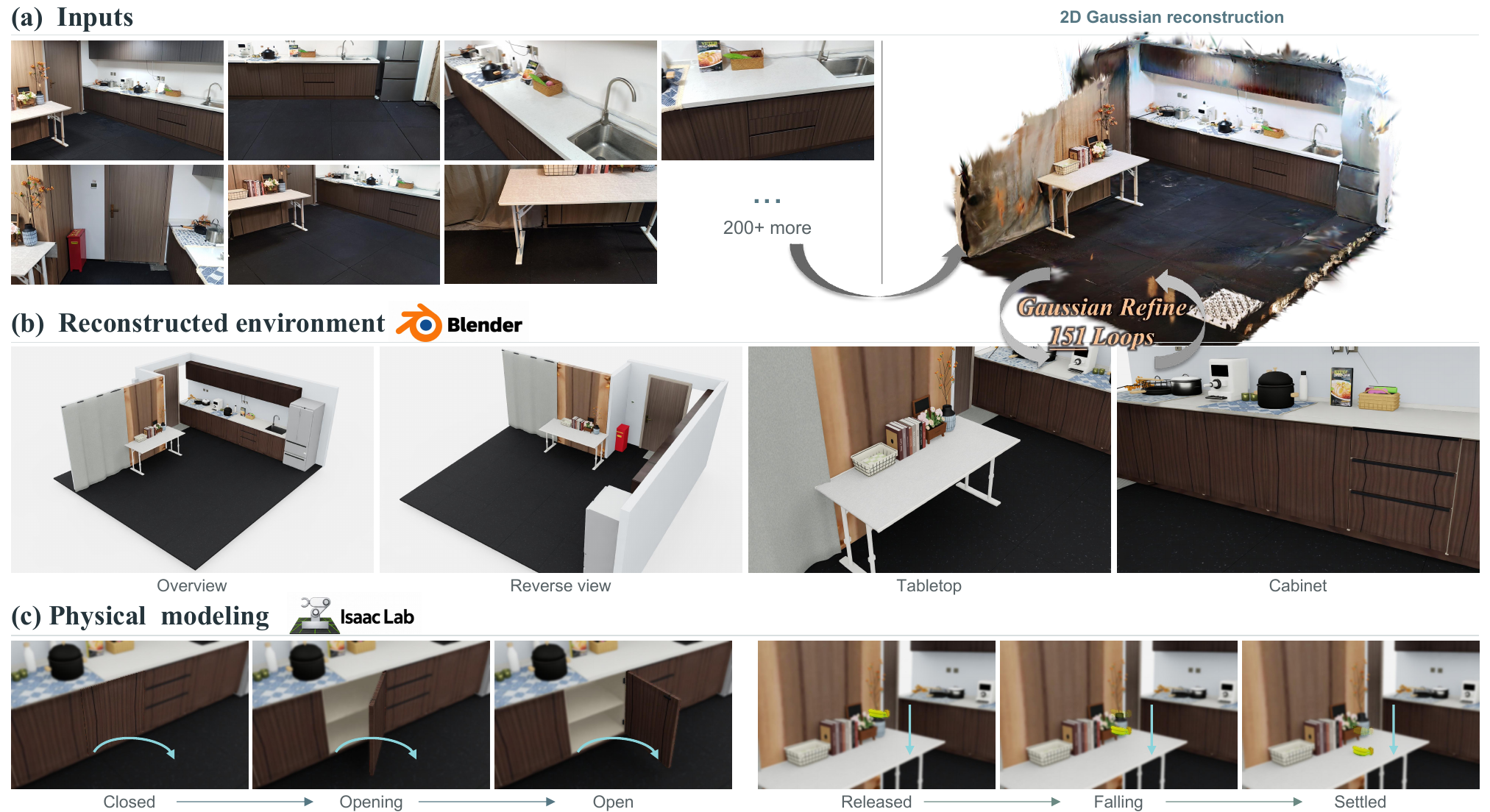}
    \captionof{figure}{\textbf{Agentic Reconstruction and Verification.}
    (a) Captured RGB views, estimated camera poses, SfM geometry, and Gaussian reconstruction supply the inputs.
    (b) Overview and close-up views illustrate mesh-scene refinement in Blender.
    (c) Cabinet opening and object-release tests check articulation and supporting contacts in IsaacLab.}
    \label{fig:agentic_real2sim_comparison}
\end{minipage}
\par\addvspace{6pt}

\paragraph{Visual Refinement and Physical Checks.}
The agent compares mesh and Gaussian renderings from matched viewpoints to refine furniture geometry and local object placement. After import into IsaacLab, cabinet-opening tests check articulated motion, while object-release tests check gravity-driven motion and contact with supporting surfaces. These checks complement the geometric comparison: visual agreement alone does not verify that a reconstructed scene supports the intended interactions.

\clearpage
\section{Experimental Details}
\label{app:experimental_details}
This section supplements the experiments in \cref{sec:experiments} with reconstruction diagnostics, baseline details, and additional results. The system setup and evaluation configurations are presented in Section~\ref{sec:exp_setup}.

\subsection{Agentic Reconstruction Evaluation}
\label{app:reconstruction_evaluation}

This section provides additional evaluation details for the agentic reconstruction described in \cref{sec:real2sim}.
We compare three reconstruction variants against a ground-truth scan captured with a handheld Q9000 LiDAR scanner: GS-Mesh, our full agentic reconstruction, and a variant without Gaussian guidance. Reconstruction accuracy is measured using symmetric Chamfer distance (CD) within the table and cabinet interaction regions, where geometric errors directly affect contact and object placement. CD is computed as the average of the two directional nearest-neighbor distances between points sampled from the reconstructed mesh and the LiDAR scan.

As reported in~\cref{tab:reconstruction_geometry}, GS-Mesh achieves a CD of 5.894\,cm, while our full reconstruction reduces it to 5.517\,cm. Without Gaussian guidance, the error increases to 11.689\,cm. This variant retains the captured images and estimated camera poses but removes Gaussian-derived mesh initialization and matched-view Gaussian rendering feedback. Qualitative comparisons in~\cref{fig:reconstruction_evaluation} further show that Gaussian guidance improves furniture dimensions, support-surface geometry, and local object placement.

\par\addvspace{6pt}
\noindent\begin{minipage}{\linewidth}
\centering
    \includegraphics[width=\linewidth]{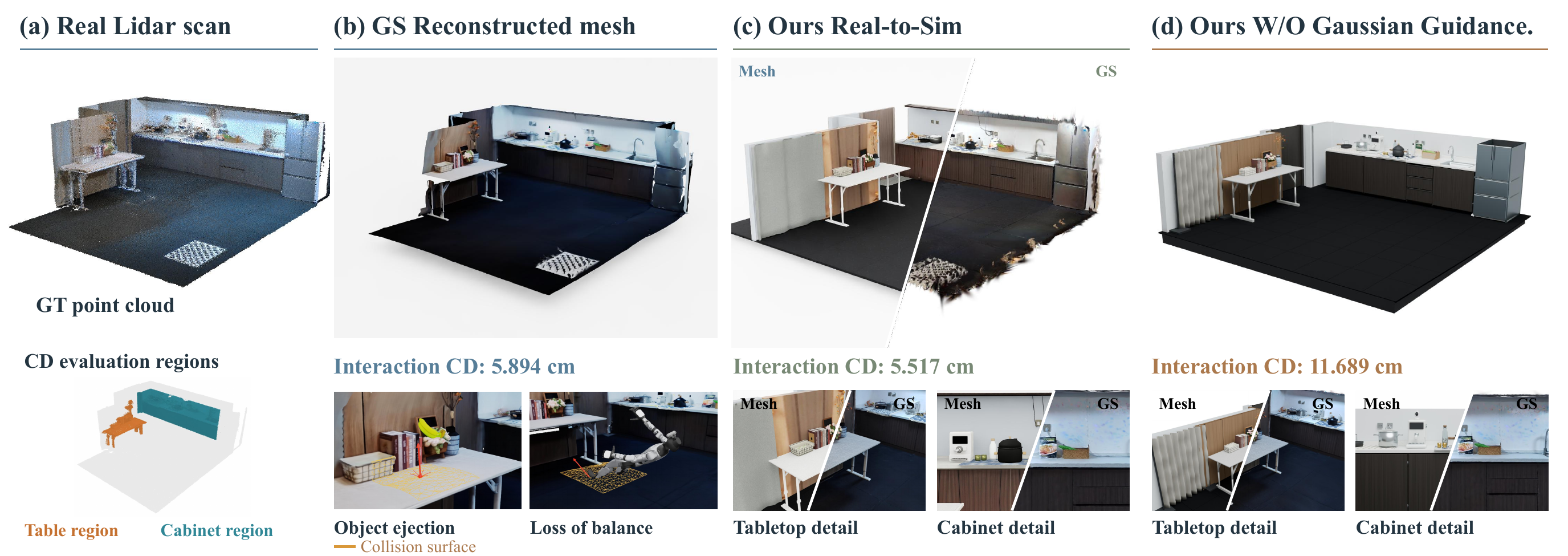}
    \captionof{figure}{\textbf{Agentic Reconstruction Comparison.}
    (a) LiDAR ground truth and interaction evaluation regions.
    (b) GS-Mesh with schematic collision examples.
    (c--d) Reconstructions with and without Gaussian guidance, with mesh--Gaussian detail views.}
    \label{fig:reconstruction_evaluation}
\end{minipage}
\par\addvspace{6pt}

\clearpage
\subsection{GPT-6 Astra Baseline}
\label{app:astra_baseline}

This section details the implementation of the Sim-Astra baseline introduced in \cref{sec:exp_setup}. GPT-6 Astra receives the same reconstructed scenes, robot and object assets, task configurations, and source demonstrations as our method, with additional access to the full simulator state. The agent autonomously designs its data-generation pipeline, either by synthesizing complete reference motions or issuing incremental closed-loop commands. It can use the SONIC planner, inverse kinematics, and whole-body control interfaces. \Cref{fig:astra_task_b_prompt} presents the complete Task B prompt, specifying the available inputs, execution constraints, success criteria, and recording requirements.

We report results from the stabilized pipeline following human-assisted debugging and task alignment. Sim-Astra generates 100 successful trajectories per task for Tasks A and B, but none for Task C. Simulator-side planning and most agent reasoning latency are excluded from the recorded demonstrations, which contain only the executed robot trajectories used for policy training.

The generated trajectories frequently exhibit pauses and repeated attempts, often exceeding one minute compared with approximately 30 seconds for real-world demonstrations. Policies trained on these data show similar hesitation during real-world deployment, suggesting that such temporal irregularities may carry over into learned behavior.

\par\addvspace{6pt}
\noindent\begin{minipage}{\linewidth}
\centering
    \begingroup
    \setlength{\fboxsep}{9pt}
    \setlength{\fboxrule}{0.4pt}
    \fcolorbox{black!25}{black!3}{%
    \begin{minipage}{\dimexpr\textwidth-2\fboxsep-2\fboxrule\relax}
    \small
    \textbf{Task B: Data Generation Prompt}\par\vspace{4pt}
    \itshape
    \setlength{\parindent}{0pt}
    \setlength{\parskip}{3pt}
    You are an autonomous agent for generating robot demonstration data. You are provided with all robot and scene assets, a physics simulation environment, the SONIC planner and whole-body control interfaces, and source demonstrations covering grasping and standing up, and object placement.

    Goal: Generate 100 successful robot manipulation trajectories. Randomize the robot's initial pose, the hat's pose, and furniture positions within the specified bounds. Imitate the source demonstrations' action sequence, whole-body poses, and motion trajectories to perform the complete task: crouch to grasp the hat from a small table, stand up, carry it to a cabinet, place it on the cabinet surface, and release it. Adapt the motions to each randomized scene while preserving the behavioral characteristics of the demonstrations.

    Generate whole-body reference motions autonomously. You may call the SONIC planner to plan or supplement reference trajectories, then execute them through the SONIC whole-body controller in physics simulation. Continuously monitor robot states, object states, and contacts, and adjust control during grasping, walking, and placement using closed-loop feedback. Analyze failures and retry until 100 successful trajectories have been collected. Do not achieve success by writing states frame by frame, teleporting objects, or rigidly attaching the hat to the hand.

    A successful trajectory must satisfy the specified acceptance criteria: the robot remains upright without prohibited collisions, the hat is not dropped during transport, and the released hat comes to rest within the designated region of the cabinet surface. Save complete executed state trajectories, timestamps, scene configurations, and source-demonstration correspondences. Record the outcome, failure reason where applicable, and generation time for every attempt.
    \end{minipage}}
    \endgroup
    \captionof{figure}{\textbf{Verbatim Task B Prompt for the GPT-6 Astra Baseline.} The prompt gives the agent the shared assets, source demonstrations, planning and control interfaces, execution restrictions, acceptance criteria, and recording requirements.}
    \label{fig:astra_task_b_prompt}
\end{minipage}
\par\addvspace{6pt}

\clearpage
\subsection{Generalization Evaluation Details}
\label{app:generalization}

This section details the five settings summarized in \cref{tab:task_a_b_full_spatial} (\cref{sec:sim2real}), using the locations in \cref{fig:ood_position}. Each configuration has five real-world trials. Entries give successes out of five; \emph{Changed} reports percentages excluding \textsc{Base}. We compare Real-Only(100) with Sim-Ours(800)$^{*}$. The Task A robot-start and tabletop-location evaluations (picking and placement) use the same set of 800 simulation demonstrations. The object--goal evaluation uses a separately generated set of 800 demonstrations, with 400 for each of the two goal types.

    \par\addvspace{10pt}\noindent\begin{minipage}[t]{\linewidth}
    \captionsetup{position=top,justification=raggedright,singlelinecheck=false,skip=3pt,belowskip=0pt}
    \centering\begin{threeparttable}
    \caption{\textbf{Task A: Robot Starts.}}
    \centering
    \label{tab:task_a_full_spatial}
    \small\setlength{\tabcolsep}{4pt}
    
    \begin{tabular}{lccccccccc}
    \toprule
      Training Data & [0,0] & [1,0] & \textsc{Base} & [3,0] & [0,1] & [1,1] & [2,1] & [3,1] & Changed (\%) \\
    \midrule
      Real-Only(100) & 2/5 & 1/5 & 4/5 & 0/5 & 0/5 & 1/5 & 4/5 & 0/5 & 22.9 \\
      \textbf{Sim-Ours(800)}$^{*}$  & 4/5 & 4/5 & 5/5 & 4/5 & 4/5 & 4/5 & 3/5 & 2/5 & \textbf{71.4} \\
    \bottomrule
    \end{tabular}
    
    \par\vspace{2pt}
    \begin{tablenotes}[flushleft]
    \footnotesize\item[] Seven \textsc{Expanded} starting locations in \cref{fig:ood_position}(a).
    \end{tablenotes}
    \end{threeparttable}
    \end{minipage}

    \par\addvspace{10pt}\noindent\begin{minipage}[t]{\linewidth}
    \captionsetup{position=top,justification=raggedright,singlelinecheck=false,skip=3pt,belowskip=0pt}
    \centering\begin{threeparttable}
    \caption{\textbf{Task B: Robot Starts.}}
    \centering
    \label{tab:task_b_full_spatial}
    \small
    
    \begin{tabular}{lcccc}
    \toprule
      Training Data & \textsc{Base} & [0,1] & [1,1] & Changed (\%) \\
    \midrule
      Real-Only(100) & 2/5 & 1/5 & 0/5 & 10.0 \\
      \textbf{Sim-Ours(800)}$^{*}$  & 4/5 & 3/5 & 4/5 & \textbf{70.0} \\
    \bottomrule
    \end{tabular}
    
    \par\vspace{2pt}
    \begin{tablenotes}[flushleft]
    \footnotesize\item[] Two \textsc{Expanded} starting locations in \cref{fig:ood_position}(b).
    \end{tablenotes}
    \end{threeparttable}
    \end{minipage}

    \par\addvspace{10pt}\noindent\begin{minipage}[t]{\linewidth}
    \captionsetup{position=top,justification=raggedright,singlelinecheck=false,skip=3pt,belowskip=0pt}
    \centering\begin{threeparttable}
    \caption{\textbf{Task A: Picking Locations.}}
    \centering
    \label{tab:task_a_pick_place_spatial}
    \small
    \setlength{\tabcolsep}{3pt}
    
    \begin{tabular}{llcccccccccc}
    \toprule
    Training Data & Stage & [0,0] & [0,1] & [0,2] & [1,0] & \textsc{Base} & [1,2] & [2,0] & [2,1] & [2,2] & Changed (\%) \\
    \midrule
    Real-Only(100) & Pick  & 1/5 & 2/5 & 2/5 & \textbf{5/5} & 1/5 & 0/5 & 3/5 & \textbf{3/5} & 0/5 & 40.0 \\
                   & Place & 0/5 & 2/5 & 0/5 & \textbf{4/5} & 1/5 & 0/5 & 3/5 & 1/5 & 0/5 & 25.0 \\
    \textbf{Sim-Ours(800)}$^{*}$ & Pick & \textbf{3/5} & \textbf{5/5} & \textbf{3/5} & 3/5 & \textbf{2/5} & \textbf{2/5} & \textbf{5/5} & 2/5 & \textbf{4/5} & \textbf{67.5} \\
                   & Place & \textbf{3/5} & \textbf{5/5} & \textbf{3/5} & 3/5 & \textbf{2/5} & \textbf{2/5} & \textbf{5/5} & \textbf{2/5} & \textbf{4/5} & \textbf{67.5} \\
    \bottomrule
    \end{tabular}
    
    \par\vspace{2pt}
    \begin{tablenotes}[flushleft]
    \footnotesize\item[] The banana varies across eight \textsc{Expanded} locations while the basket remains fixed (\cref{fig:ood_position}(c)).
    \end{tablenotes}
    \end{threeparttable}
    \end{minipage}

    \par\addvspace{10pt}\noindent\begin{minipage}[t]{\linewidth}
    \captionsetup{position=top,justification=raggedright,singlelinecheck=false,skip=3pt,belowskip=0pt}
    \centering\begin{threeparttable}
    \caption{\textbf{Task A: Placement Locations.}}
    \centering
    \label{tab:task_a_placement_spatial}
    \small
    \setlength{\tabcolsep}{8pt}
 
    \begin{tabular}{l cc cc cc cc}
    \toprule
     & \multicolumn{2}{c}{[0,0]} & \multicolumn{2}{c}{\textsc{Base}} & \multicolumn{2}{c}{[0,1]} & \multicolumn{2}{c}{Changed (\%)} \\
    Training Data & Pick & Place & Pick & Place & Pick & Place & Pick & Place \\
    \midrule
     Real-Only(100) & \textbf{4/5} & 0/5 & \textbf{2/5} & \textbf{2/5} & 3/5 & \textbf{3/5} & \textbf{70.0} & 30.0 \\
     \textbf{Sim-Ours(800)}$^{*}$ & 2/5 & \textbf{2/5} & \textbf{2/5} & \textbf{2/5} & \textbf{5/5} & \textbf{3/5} & \textbf{70.0} & \textbf{50.0} \\
    \bottomrule
    \end{tabular}
    
    \par\vspace{2pt}
    \begin{tablenotes}[flushleft]
    \footnotesize\item[] The basket varies across two \textsc{Expanded} locations while the banana remains fixed (\cref{fig:ood_position}(c)).
    \end{tablenotes}
    \end{threeparttable}
    \end{minipage}

    \par\addvspace{10pt}\noindent\begin{minipage}[t]{\linewidth}
    \captionsetup{position=top,justification=raggedright,singlelinecheck=false,skip=3pt,belowskip=0pt}
    \centering\begin{threeparttable}
    \caption{\textbf{Task A: Object--Goal Pair.}}
    \centering
    \label{tab:object_goal_generalization}
    \small
        \begin{tabular}{lccc}
    \toprule
      Training Data & Banana to Basket (\textsc{Base}) & Bitter Melon to White Plate & Changed (\%) \\
    \midrule
      Real-Only(100) & 2/5 & 1/5 & 20.0 \\
      \textbf{Sim-Ours(800)}$^{*}$ & 5/5 & 5/5 & 100.0 \\
    \bottomrule
    \end{tabular}
    \par\vspace{2pt}
    \begin{tablenotes}[flushleft]
    \footnotesize\item[] The alternative pair replaces the banana and basket with bitter melon and a white plate.
    \end{tablenotes}
    \end{threeparttable}
    \end{minipage}

Pooling the five \textsc{Expanded} settings gives 100 trials per method (35, 10, 40, 10, and 5 by setting). Real-Only(100) and Sim-Ours(800)$^{*}$ achieve 23/100 and 69/100 successes, matching \cref{tab:task_a_b_full_spatial}.

\clearpage
\section{Additional Qualitative Results}
\label{app:qualitative_results}
\label{app:more_visualization}

\Cref{fig:more_visualization_page_one} visualizes generated simulation demonstrations for Tasks A--C. Each example pairs third-person views of a simulated rollout with the corresponding rendered egocentric observations used for policy training.

\par\addvspace{6pt}
\noindent\begin{minipage}{\linewidth}
\centering
    \includegraphics[width=\textwidth,height=0.80\textheight,keepaspectratio]{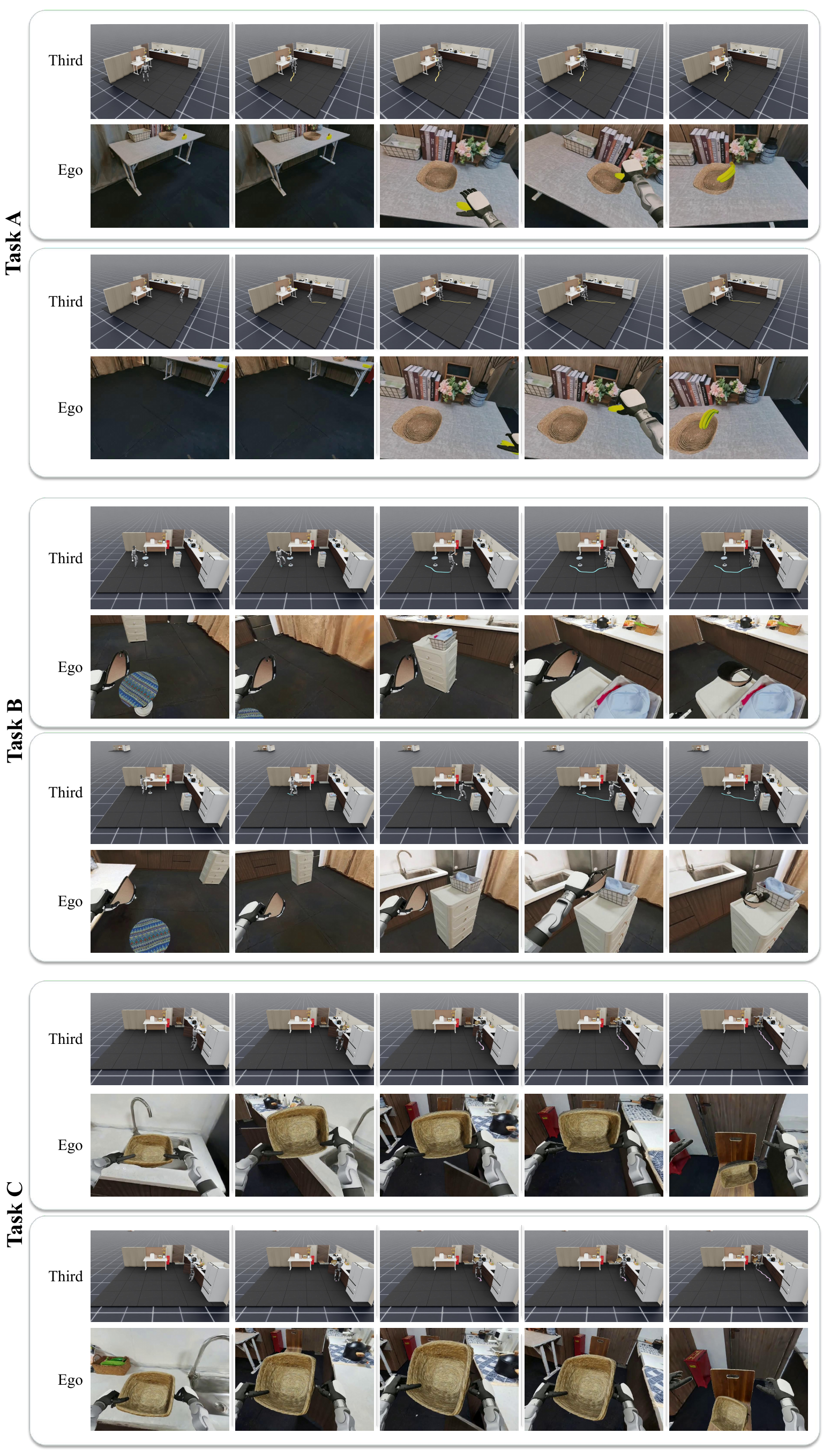}
    \captionof{figure}{\textbf{Generated Simulation Demonstrations.}
    Third-person simulation views and corresponding egocentric renderings illustrate locomotion, manipulation, and leg-assisted interactions across Tasks A--C.}
    \label{fig:more_visualization_page_one}
\end{minipage}
\par\addvspace{6pt}

\endgroup

\end{document}